\documentclass[conference]{IEEEtran}
\IEEEoverridecommandlockouts
\usepackage{mathptmx} 
\usepackage{cite}
\usepackage{amsmath,amssymb,amsfonts}
\usepackage{algorithmic}
\usepackage{textcomp}
\usepackage{xcolor}

\def\BibTeX{{\rm B\kern-.05em{\sc i\kern-.025em b}\kern-.08em
    T\kern-.1667em\lower.7ex\hbox{E}\kern-.125emX}}

\newcommand{\ignore}[1]{}
\usepackage[pass]{geometry}
\usepackage{fancyhdr}
\usepackage[normalem]{ulem}
\usepackage[hyphens]{url}
\usepackage{color}
\usepackage{soul}
\usepackage{microtype}

\usepackage{lipsum}
\usepackage{subfig}
\usepackage{graphicx}
\usepackage{listings}
\usepackage{enumitem}
\usepackage{multirow}
\usepackage{hhline}
\usepackage{tabularx}
\usepackage{makecell}
\usepackage[draft]{minted}

\usepackage{siunitx}
\usepackage{pifont}
\usepackage[layout={inline}]{fixme}
\usepackage[font=small,skip=12pt]{caption}

\makeatletter
\def\PYGdefault@reset{\let\PYGdefault@it=\relax \let\PYGdefault@bf=\relax%
    \let\PYGdefault@ul=\relax \let\PYGdefault@tc=\relax%
    \let\PYGdefault@bc=\relax \let\PYGdefault@ff=\relax}
\def\PYGdefault@tok#1{\csname PYGdefault@tok@#1\endcsname}
\def\PYGdefault@toks#1+{\ifx\relax#1\empty\else%
    \PYGdefault@tok{#1}\expandafter\PYGdefault@toks\fi}
\def\PYGdefault@do#1{\PYGdefault@bc{\PYGdefault@tc{\PYGdefault@ul{%
    \PYGdefault@it{\PYGdefault@bf{\PYGdefault@ff{#1}}}}}}}
\def\PYGdefault#1#2{\PYGdefault@reset\PYGdefault@toks#1+\relax+\PYGdefault@do{#2}}

\expandafter\def\csname PYGdefault@tok@gd\endcsname{\def\PYGdefault@tc##1{\textcolor[rgb]{0.63,0.00,0.00}{##1}}}
\expandafter\def\csname PYGdefault@tok@gu\endcsname{\let\PYGdefault@bf=\textbf\def\PYGdefault@tc##1{\textcolor[rgb]{0.50,0.00,0.50}{##1}}}
\expandafter\def\csname PYGdefault@tok@gt\endcsname{\def\PYGdefault@tc##1{\textcolor[rgb]{0.00,0.27,0.87}{##1}}}
\expandafter\def\csname PYGdefault@tok@gs\endcsname{\let\PYGdefault@bf=\textbf}
\expandafter\def\csname PYGdefault@tok@gr\endcsname{\def\PYGdefault@tc##1{\textcolor[rgb]{1.00,0.00,0.00}{##1}}}
\expandafter\def\csname PYGdefault@tok@cm\endcsname{\let\PYGdefault@it=\textit\def\PYGdefault@tc##1{\textcolor[rgb]{0.25,0.50,0.50}{##1}}}
\expandafter\def\csname PYGdefault@tok@vg\endcsname{\def\PYGdefault@tc##1{\textcolor[rgb]{0.10,0.09,0.49}{##1}}}
\expandafter\def\csname PYGdefault@tok@vi\endcsname{\def\PYGdefault@tc##1{\textcolor[rgb]{0.10,0.09,0.49}{##1}}}
\expandafter\def\csname PYGdefault@tok@vm\endcsname{\def\PYGdefault@tc##1{\textcolor[rgb]{0.10,0.09,0.49}{##1}}}
\expandafter\def\csname PYGdefault@tok@mh\endcsname{\def\PYGdefault@tc##1{\textcolor[rgb]{0.40,0.40,0.40}{##1}}}
\expandafter\def\csname PYGdefault@tok@cs\endcsname{\let\PYGdefault@it=\textit\def\PYGdefault@tc##1{\textcolor[rgb]{0.25,0.50,0.50}{##1}}}
\expandafter\def\csname PYGdefault@tok@ge\endcsname{\let\PYGdefault@it=\textit}
\expandafter\def\csname PYGdefault@tok@vc\endcsname{\def\PYGdefault@tc##1{\textcolor[rgb]{0.10,0.09,0.49}{##1}}}
\expandafter\def\csname PYGdefault@tok@il\endcsname{\def\PYGdefault@tc##1{\textcolor[rgb]{0.40,0.40,0.40}{##1}}}
\expandafter\def\csname PYGdefault@tok@go\endcsname{\def\PYGdefault@tc##1{\textcolor[rgb]{0.53,0.53,0.53}{##1}}}
\expandafter\def\csname PYGdefault@tok@cp\endcsname{\def\PYGdefault@tc##1{\textcolor[rgb]{0.74,0.48,0.00}{##1}}}
\expandafter\def\csname PYGdefault@tok@gi\endcsname{\def\PYGdefault@tc##1{\textcolor[rgb]{0.00,0.63,0.00}{##1}}}
\expandafter\def\csname PYGdefault@tok@gh\endcsname{\let\PYGdefault@bf=\textbf\def\PYGdefault@tc##1{\textcolor[rgb]{0.00,0.00,0.50}{##1}}}
\expandafter\def\csname PYGdefault@tok@ni\endcsname{\let\PYGdefault@bf=\textbf\def\PYGdefault@tc##1{\textcolor[rgb]{0.60,0.60,0.60}{##1}}}
\expandafter\def\csname PYGdefault@tok@nl\endcsname{\def\PYGdefault@tc##1{\textcolor[rgb]{0.63,0.63,0.00}{##1}}}
\expandafter\def\csname PYGdefault@tok@nn\endcsname{\let\PYGdefault@bf=\textbf\def\PYGdefault@tc##1{\textcolor[rgb]{0.00,0.00,1.00}{##1}}}
\expandafter\def\csname PYGdefault@tok@no\endcsname{\def\PYGdefault@tc##1{\textcolor[rgb]{0.53,0.00,0.00}{##1}}}
\expandafter\def\csname PYGdefault@tok@na\endcsname{\def\PYGdefault@tc##1{\textcolor[rgb]{0.49,0.56,0.16}{##1}}}
\expandafter\def\csname PYGdefault@tok@nb\endcsname{\def\PYGdefault@tc##1{\textcolor[rgb]{0.00,0.50,0.00}{##1}}}
\expandafter\def\csname PYGdefault@tok@nc\endcsname{\let\PYGdefault@bf=\textbf\def\PYGdefault@tc##1{\textcolor[rgb]{0.00,0.00,1.00}{##1}}}
\expandafter\def\csname PYGdefault@tok@nd\endcsname{\def\PYGdefault@tc##1{\textcolor[rgb]{0.67,0.13,1.00}{##1}}}
\expandafter\def\csname PYGdefault@tok@ne\endcsname{\let\PYGdefault@bf=\textbf\def\PYGdefault@tc##1{\textcolor[rgb]{0.82,0.25,0.23}{##1}}}
\expandafter\def\csname PYGdefault@tok@nf\endcsname{\def\PYGdefault@tc##1{\textcolor[rgb]{0.00,0.00,1.00}{##1}}}
\expandafter\def\csname PYGdefault@tok@si\endcsname{\let\PYGdefault@bf=\textbf\def\PYGdefault@tc##1{\textcolor[rgb]{0.73,0.40,0.53}{##1}}}
\expandafter\def\csname PYGdefault@tok@s2\endcsname{\def\PYGdefault@tc##1{\textcolor[rgb]{0.73,0.13,0.13}{##1}}}
\expandafter\def\csname PYGdefault@tok@nt\endcsname{\let\PYGdefault@bf=\textbf\def\PYGdefault@tc##1{\textcolor[rgb]{0.00,0.50,0.00}{##1}}}
\expandafter\def\csname PYGdefault@tok@nv\endcsname{\def\PYGdefault@tc##1{\textcolor[rgb]{0.10,0.09,0.49}{##1}}}
\expandafter\def\csname PYGdefault@tok@s1\endcsname{\def\PYGdefault@tc##1{\textcolor[rgb]{0.73,0.13,0.13}{##1}}}
\expandafter\def\csname PYGdefault@tok@dl\endcsname{\def\PYGdefault@tc##1{\textcolor[rgb]{0.73,0.13,0.13}{##1}}}
\expandafter\def\csname PYGdefault@tok@ch\endcsname{\let\PYGdefault@it=\textit\def\PYGdefault@tc##1{\textcolor[rgb]{0.25,0.50,0.50}{##1}}}
\expandafter\def\csname PYGdefault@tok@m\endcsname{\def\PYGdefault@tc##1{\textcolor[rgb]{0.40,0.40,0.40}{##1}}}
\expandafter\def\csname PYGdefault@tok@gp\endcsname{\let\PYGdefault@bf=\textbf\def\PYGdefault@tc##1{\textcolor[rgb]{0.00,0.00,0.50}{##1}}}
\expandafter\def\csname PYGdefault@tok@sh\endcsname{\def\PYGdefault@tc##1{\textcolor[rgb]{0.73,0.13,0.13}{##1}}}
\expandafter\def\csname PYGdefault@tok@ow\endcsname{\let\PYGdefault@bf=\textbf\def\PYGdefault@tc##1{\textcolor[rgb]{0.67,0.13,1.00}{##1}}}
\expandafter\def\csname PYGdefault@tok@sx\endcsname{\def\PYGdefault@tc##1{\textcolor[rgb]{0.00,0.50,0.00}{##1}}}
\expandafter\def\csname PYGdefault@tok@bp\endcsname{\def\PYGdefault@tc##1{\textcolor[rgb]{0.00,0.50,0.00}{##1}}}
\expandafter\def\csname PYGdefault@tok@c1\endcsname{\let\PYGdefault@it=\textit\def\PYGdefault@tc##1{\textcolor[rgb]{0.25,0.50,0.50}{##1}}}
\expandafter\def\csname PYGdefault@tok@fm\endcsname{\def\PYGdefault@tc##1{\textcolor[rgb]{0.00,0.00,1.00}{##1}}}
\expandafter\def\csname PYGdefault@tok@o\endcsname{\def\PYGdefault@tc##1{\textcolor[rgb]{0.40,0.40,0.40}{##1}}}
\expandafter\def\csname PYGdefault@tok@kc\endcsname{\let\PYGdefault@bf=\textbf\def\PYGdefault@tc##1{\textcolor[rgb]{0.00,0.50,0.00}{##1}}}
\expandafter\def\csname PYGdefault@tok@c\endcsname{\let\PYGdefault@it=\textit\def\PYGdefault@tc##1{\textcolor[rgb]{0.25,0.50,0.50}{##1}}}
\expandafter\def\csname PYGdefault@tok@mf\endcsname{\def\PYGdefault@tc##1{\textcolor[rgb]{0.40,0.40,0.40}{##1}}}
\expandafter\def\csname PYGdefault@tok@err\endcsname{\def\PYGdefault@bc##1{\setlength{\fboxsep}{0pt}\fcolorbox[rgb]{1.00,0.00,0.00}{1,1,1}{\strut ##1}}}
\expandafter\def\csname PYGdefault@tok@mb\endcsname{\def\PYGdefault@tc##1{\textcolor[rgb]{0.40,0.40,0.40}{##1}}}
\expandafter\def\csname PYGdefault@tok@ss\endcsname{\def\PYGdefault@tc##1{\textcolor[rgb]{0.10,0.09,0.49}{##1}}}
\expandafter\def\csname PYGdefault@tok@sr\endcsname{\def\PYGdefault@tc##1{\textcolor[rgb]{0.73,0.40,0.53}{##1}}}
\expandafter\def\csname PYGdefault@tok@mo\endcsname{\def\PYGdefault@tc##1{\textcolor[rgb]{0.40,0.40,0.40}{##1}}}
\expandafter\def\csname PYGdefault@tok@kd\endcsname{\let\PYGdefault@bf=\textbf\def\PYGdefault@tc##1{\textcolor[rgb]{0.00,0.50,0.00}{##1}}}
\expandafter\def\csname PYGdefault@tok@mi\endcsname{\def\PYGdefault@tc##1{\textcolor[rgb]{0.40,0.40,0.40}{##1}}}
\expandafter\def\csname PYGdefault@tok@kn\endcsname{\let\PYGdefault@bf=\textbf\def\PYGdefault@tc##1{\textcolor[rgb]{0.00,0.50,0.00}{##1}}}
\expandafter\def\csname PYGdefault@tok@cpf\endcsname{\let\PYGdefault@it=\textit\def\PYGdefault@tc##1{\textcolor[rgb]{0.25,0.50,0.50}{##1}}}
\expandafter\def\csname PYGdefault@tok@kr\endcsname{\let\PYGdefault@bf=\textbf\def\PYGdefault@tc##1{\textcolor[rgb]{0.00,0.50,0.00}{##1}}}
\expandafter\def\csname PYGdefault@tok@s\endcsname{\def\PYGdefault@tc##1{\textcolor[rgb]{0.73,0.13,0.13}{##1}}}
\expandafter\def\csname PYGdefault@tok@kp\endcsname{\def\PYGdefault@tc##1{\textcolor[rgb]{0.00,0.50,0.00}{##1}}}
\expandafter\def\csname PYGdefault@tok@w\endcsname{\def\PYGdefault@tc##1{\textcolor[rgb]{0.73,0.73,0.73}{##1}}}
\expandafter\def\csname PYGdefault@tok@kt\endcsname{\def\PYGdefault@tc##1{\textcolor[rgb]{0.69,0.00,0.25}{##1}}}
\expandafter\def\csname PYGdefault@tok@sc\endcsname{\def\PYGdefault@tc##1{\textcolor[rgb]{0.73,0.13,0.13}{##1}}}
\expandafter\def\csname PYGdefault@tok@sb\endcsname{\def\PYGdefault@tc##1{\textcolor[rgb]{0.73,0.13,0.13}{##1}}}
\expandafter\def\csname PYGdefault@tok@sa\endcsname{\def\PYGdefault@tc##1{\textcolor[rgb]{0.73,0.13,0.13}{##1}}}
\expandafter\def\csname PYGdefault@tok@k\endcsname{\let\PYGdefault@bf=\textbf\def\PYGdefault@tc##1{\textcolor[rgb]{0.00,0.50,0.00}{##1}}}
\expandafter\def\csname PYGdefault@tok@se\endcsname{\let\PYGdefault@bf=\textbf\def\PYGdefault@tc##1{\textcolor[rgb]{0.73,0.40,0.13}{##1}}}
\expandafter\def\csname PYGdefault@tok@sd\endcsname{\let\PYGdefault@it=\textit\def\PYGdefault@tc##1{\textcolor[rgb]{0.73,0.13,0.13}{##1}}}

\makeatother

\makeatletter
\def\PYG@reset{\let\PYG@it=\relax \let\PYG@bf=\relax%
    \let\PYG@ul=\relax \let\PYG@tc=\relax%
    \let\PYG@bc=\relax \let\PYG@ff=\relax}
\def\PYG@tok#1{\csname PYG@tok@#1\endcsname}
\def\PYG@toks#1+{\ifx\relax#1\empty\else%
    \PYG@tok{#1}\expandafter\PYG@toks\fi}
\def\PYG@do#1{\PYG@bc{\PYG@tc{\PYG@ul{%
    \PYG@it{\PYG@bf{\PYG@ff{#1}}}}}}}
\def\PYG#1#2{\PYG@reset\PYG@toks#1+\relax+\PYG@do{#2}}

\expandafter\def\csname PYG@tok@gd\endcsname{\def\PYG@tc##1{\textcolor[rgb]{0.63,0.00,0.00}{##1}}}
\expandafter\def\csname PYG@tok@gu\endcsname{\let\PYG@bf=\textbf\def\PYG@tc##1{\textcolor[rgb]{0.50,0.00,0.50}{##1}}}
\expandafter\def\csname PYG@tok@gt\endcsname{\def\PYG@tc##1{\textcolor[rgb]{0.00,0.27,0.87}{##1}}}
\expandafter\def\csname PYG@tok@gs\endcsname{\let\PYG@bf=\textbf}
\expandafter\def\csname PYG@tok@gr\endcsname{\def\PYG@tc##1{\textcolor[rgb]{1.00,0.00,0.00}{##1}}}
\expandafter\def\csname PYG@tok@cm\endcsname{\let\PYG@it=\textit\def\PYG@tc##1{\textcolor[rgb]{0.25,0.50,0.50}{##1}}}
\expandafter\def\csname PYG@tok@vg\endcsname{\def\PYG@tc##1{\textcolor[rgb]{0.10,0.09,0.49}{##1}}}
\expandafter\def\csname PYG@tok@vi\endcsname{\def\PYG@tc##1{\textcolor[rgb]{0.10,0.09,0.49}{##1}}}
\expandafter\def\csname PYG@tok@vm\endcsname{\def\PYG@tc##1{\textcolor[rgb]{0.10,0.09,0.49}{##1}}}
\expandafter\def\csname PYG@tok@mh\endcsname{\def\PYG@tc##1{\textcolor[rgb]{0.40,0.40,0.40}{##1}}}
\expandafter\def\csname PYG@tok@cs\endcsname{\let\PYG@it=\textit\def\PYG@tc##1{\textcolor[rgb]{0.25,0.50,0.50}{##1}}}
\expandafter\def\csname PYG@tok@ge\endcsname{\let\PYG@it=\textit}
\expandafter\def\csname PYG@tok@vc\endcsname{\def\PYG@tc##1{\textcolor[rgb]{0.10,0.09,0.49}{##1}}}
\expandafter\def\csname PYG@tok@il\endcsname{\def\PYG@tc##1{\textcolor[rgb]{0.40,0.40,0.40}{##1}}}
\expandafter\def\csname PYG@tok@go\endcsname{\def\PYG@tc##1{\textcolor[rgb]{0.53,0.53,0.53}{##1}}}
\expandafter\def\csname PYG@tok@cp\endcsname{\def\PYG@tc##1{\textcolor[rgb]{0.74,0.48,0.00}{##1}}}
\expandafter\def\csname PYG@tok@gi\endcsname{\def\PYG@tc##1{\textcolor[rgb]{0.00,0.63,0.00}{##1}}}
\expandafter\def\csname PYG@tok@gh\endcsname{\let\PYG@bf=\textbf\def\PYG@tc##1{\textcolor[rgb]{0.00,0.00,0.50}{##1}}}
\expandafter\def\csname PYG@tok@ni\endcsname{\let\PYG@bf=\textbf\def\PYG@tc##1{\textcolor[rgb]{0.60,0.60,0.60}{##1}}}
\expandafter\def\csname PYG@tok@nl\endcsname{\def\PYG@tc##1{\textcolor[rgb]{0.63,0.63,0.00}{##1}}}
\expandafter\def\csname PYG@tok@nn\endcsname{\let\PYG@bf=\textbf\def\PYG@tc##1{\textcolor[rgb]{0.00,0.00,1.00}{##1}}}
\expandafter\def\csname PYG@tok@no\endcsname{\def\PYG@tc##1{\textcolor[rgb]{0.53,0.00,0.00}{##1}}}
\expandafter\def\csname PYG@tok@na\endcsname{\def\PYG@tc##1{\textcolor[rgb]{0.49,0.56,0.16}{##1}}}
\expandafter\def\csname PYG@tok@nb\endcsname{\def\PYG@tc##1{\textcolor[rgb]{0.00,0.50,0.00}{##1}}}
\expandafter\def\csname PYG@tok@nc\endcsname{\let\PYG@bf=\textbf\def\PYG@tc##1{\textcolor[rgb]{0.00,0.00,1.00}{##1}}}
\expandafter\def\csname PYG@tok@nd\endcsname{\def\PYG@tc##1{\textcolor[rgb]{0.67,0.13,1.00}{##1}}}
\expandafter\def\csname PYG@tok@ne\endcsname{\let\PYG@bf=\textbf\def\PYG@tc##1{\textcolor[rgb]{0.82,0.25,0.23}{##1}}}
\expandafter\def\csname PYG@tok@nf\endcsname{\def\PYG@tc##1{\textcolor[rgb]{0.00,0.00,1.00}{##1}}}
\expandafter\def\csname PYG@tok@si\endcsname{\let\PYG@bf=\textbf\def\PYG@tc##1{\textcolor[rgb]{0.73,0.40,0.53}{##1}}}
\expandafter\def\csname PYG@tok@s2\endcsname{\def\PYG@tc##1{\textcolor[rgb]{0.73,0.13,0.13}{##1}}}
\expandafter\def\csname PYG@tok@nt\endcsname{\let\PYG@bf=\textbf\def\PYG@tc##1{\textcolor[rgb]{0.00,0.50,0.00}{##1}}}
\expandafter\def\csname PYG@tok@nv\endcsname{\def\PYG@tc##1{\textcolor[rgb]{0.10,0.09,0.49}{##1}}}
\expandafter\def\csname PYG@tok@s1\endcsname{\def\PYG@tc##1{\textcolor[rgb]{0.73,0.13,0.13}{##1}}}
\expandafter\def\csname PYG@tok@dl\endcsname{\def\PYG@tc##1{\textcolor[rgb]{0.73,0.13,0.13}{##1}}}
\expandafter\def\csname PYG@tok@ch\endcsname{\let\PYG@it=\textit\def\PYG@tc##1{\textcolor[rgb]{0.25,0.50,0.50}{##1}}}
\expandafter\def\csname PYG@tok@m\endcsname{\def\PYG@tc##1{\textcolor[rgb]{0.40,0.40,0.40}{##1}}}
\expandafter\def\csname PYG@tok@gp\endcsname{\let\PYG@bf=\textbf\def\PYG@tc##1{\textcolor[rgb]{0.00,0.00,0.50}{##1}}}
\expandafter\def\csname PYG@tok@sh\endcsname{\def\PYG@tc##1{\textcolor[rgb]{0.73,0.13,0.13}{##1}}}
\expandafter\def\csname PYG@tok@ow\endcsname{\let\PYG@bf=\textbf\def\PYG@tc##1{\textcolor[rgb]{0.67,0.13,1.00}{##1}}}
\expandafter\def\csname PYG@tok@sx\endcsname{\def\PYG@tc##1{\textcolor[rgb]{0.00,0.50,0.00}{##1}}}
\expandafter\def\csname PYG@tok@bp\endcsname{\def\PYG@tc##1{\textcolor[rgb]{0.00,0.50,0.00}{##1}}}
\expandafter\def\csname PYG@tok@c1\endcsname{\let\PYG@it=\textit\def\PYG@tc##1{\textcolor[rgb]{0.25,0.50,0.50}{##1}}}
\expandafter\def\csname PYG@tok@fm\endcsname{\def\PYG@tc##1{\textcolor[rgb]{0.00,0.00,1.00}{##1}}}
\expandafter\def\csname PYG@tok@o\endcsname{\def\PYG@tc##1{\textcolor[rgb]{0.40,0.40,0.40}{##1}}}
\expandafter\def\csname PYG@tok@kc\endcsname{\let\PYG@bf=\textbf\def\PYG@tc##1{\textcolor[rgb]{0.00,0.50,0.00}{##1}}}
\expandafter\def\csname PYG@tok@c\endcsname{\let\PYG@it=\textit\def\PYG@tc##1{\textcolor[rgb]{0.25,0.50,0.50}{##1}}}
\expandafter\def\csname PYG@tok@mf\endcsname{\def\PYG@tc##1{\textcolor[rgb]{0.40,0.40,0.40}{##1}}}
\expandafter\def\csname PYG@tok@err\endcsname{\def\PYG@bc##1{\setlength{\fboxsep}{0pt}\fcolorbox[rgb]{1.00,0.00,0.00}{1,1,1}{\strut ##1}}}
\expandafter\def\csname PYG@tok@mb\endcsname{\def\PYG@tc##1{\textcolor[rgb]{0.40,0.40,0.40}{##1}}}
\expandafter\def\csname PYG@tok@ss\endcsname{\def\PYG@tc##1{\textcolor[rgb]{0.10,0.09,0.49}{##1}}}
\expandafter\def\csname PYG@tok@sr\endcsname{\def\PYG@tc##1{\textcolor[rgb]{0.73,0.40,0.53}{##1}}}
\expandafter\def\csname PYG@tok@mo\endcsname{\def\PYG@tc##1{\textcolor[rgb]{0.40,0.40,0.40}{##1}}}
\expandafter\def\csname PYG@tok@kd\endcsname{\let\PYG@bf=\textbf\def\PYG@tc##1{\textcolor[rgb]{0.00,0.50,0.00}{##1}}}
\expandafter\def\csname PYG@tok@mi\endcsname{\def\PYG@tc##1{\textcolor[rgb]{0.40,0.40,0.40}{##1}}}
\expandafter\def\csname PYG@tok@kn\endcsname{\let\PYG@bf=\textbf\def\PYG@tc##1{\textcolor[rgb]{0.00,0.50,0.00}{##1}}}
\expandafter\def\csname PYG@tok@cpf\endcsname{\let\PYG@it=\textit\def\PYG@tc##1{\textcolor[rgb]{0.25,0.50,0.50}{##1}}}
\expandafter\def\csname PYG@tok@kr\endcsname{\let\PYG@bf=\textbf\def\PYG@tc##1{\textcolor[rgb]{0.00,0.50,0.00}{##1}}}
\expandafter\def\csname PYG@tok@s\endcsname{\def\PYG@tc##1{\textcolor[rgb]{0.73,0.13,0.13}{##1}}}
\expandafter\def\csname PYG@tok@kp\endcsname{\def\PYG@tc##1{\textcolor[rgb]{0.00,0.50,0.00}{##1}}}
\expandafter\def\csname PYG@tok@w\endcsname{\def\PYG@tc##1{\textcolor[rgb]{0.73,0.73,0.73}{##1}}}
\expandafter\def\csname PYG@tok@kt\endcsname{\def\PYG@tc##1{\textcolor[rgb]{0.69,0.00,0.25}{##1}}}
\expandafter\def\csname PYG@tok@sc\endcsname{\def\PYG@tc##1{\textcolor[rgb]{0.73,0.13,0.13}{##1}}}
\expandafter\def\csname PYG@tok@sb\endcsname{\def\PYG@tc##1{\textcolor[rgb]{0.73,0.13,0.13}{##1}}}
\expandafter\def\csname PYG@tok@sa\endcsname{\def\PYG@tc##1{\textcolor[rgb]{0.73,0.13,0.13}{##1}}}
\expandafter\def\csname PYG@tok@k\endcsname{\let\PYG@bf=\textbf\def\PYG@tc##1{\textcolor[rgb]{0.00,0.50,0.00}{##1}}}
\expandafter\def\csname PYG@tok@se\endcsname{\let\PYG@bf=\textbf\def\PYG@tc##1{\textcolor[rgb]{0.73,0.40,0.13}{##1}}}
\expandafter\def\csname PYG@tok@sd\endcsname{\let\PYG@it=\textit\def\PYG@tc##1{\textcolor[rgb]{0.73,0.13,0.13}{##1}}}

\def\PYGZus{\char`\_}
\def\PYGZob{\char`\{}
\def\PYGZcb{\char`\}}

\def\PYGZlt{\char`\<}

\def\PYGZsh{\char`\#}

\def\PYGZhy{\char`\-}
\def\PYGZsq{\char`\'}
\def\PYGZdq{\char`\"}

\makeatother

\fxsetup{theme=color, status=draft}
\let\oldding\ding
\renewcommand{\ding}[2][1]{\scalebox{#1}{\oldding{#2}}}

\newcommand*\colourcheck[1]{%
  \expandafter\newcommand\csname #1check\endcsname{\textcolor{#1}{\ding[1.4]{52}}}%
}
\newcommand*\colourcross[1]{%
  \expandafter\newcommand\csname #1cross\endcsname{\textcolor{#1}{\ding[1.4]{55}}}%
}
\colourcheck{green}
\colourcross{red}

\soulregister\ref7
\soulregister\cite7

\usepackage{mathtools}

\makeatletter
\def\footnoterule{\relax%
  \kern-5pt
  \hbox to \columnwidth{\hfill\vrule width 1\columnwidth height 0.4pt\hfill}
  \kern4.6pt}
\makeatother

\usepackage[hidelinks]{hyperref}

\newcommand{\iscasubmissionnumber}{127}

\fancypagestyle{firstpage}{
  \fancyhf{}

  \fancyhead[C]{\normalsize{ISCA 2020 Submission
      \textbf{\127\iscasubmissionnumber} \\ Confidential Draft: DO NOT DISTRIBUTE}} 
  \fancyfoot[C]{\thepage}
}

\title{SMAUG: End-to-End Full-Stack Simulation Infrastructure for Deep Learning Workloads}
\author{
Sam (Likun) Xi\IEEEauthorrefmark{1}\thanks{\IEEEauthorrefmark{1}These authors contributed equally to this work.},
Yuan Yao\IEEEauthorrefmark{1},
Kshitij Bhardwaj,
Paul Whatmough,
Gu-Yeon Wei and
David Brooks \\
\IEEEauthorblockA{Harvard University}
\IEEEauthorblockA{
slxi1202@gmail.com \hspace{0.1cm} $\{$yuanyao,kbhardwaj,pwhatmough$\}$@seas.harvard.edu \hspace{0.1cm}
$\{$gywei,dbrooks$\}$@eecs.harvard.edu}
}

\begin{document}
\maketitle
\pagestyle{plain}
\begin{abstract}
\hspace{0pt} In recent years, there has been tremendous advances in hardware acceleration of deep neural networks. However,
most of the research has focused on optimizing accelerator microarchitecture for higher performance and
energy efficiency on a per-layer basis. We find that for overall single-batch inference latency, the accelerator
may only make up 25-40\%, with the rest spent on data movement and in the deep learning
software framework. Thus far, it has been very difficult to study end-to-end DNN performance during early stage
design (before RTL is available) because there are no existing DNN frameworks that support end-to-end simulation with 
easy custom hardware accelerator integration. 
To address this gap in research infrastructure, we present SMAUG, the first DNN framework that is 
purpose-built
for simulation of end-to-end deep learning applications. SMAUG offers researchers a wide range of capabilities
for evaluating DNN workloads, from diverse network topologies to easy accelerator modeling and SoC integration.
To demonstrate the power and value of SMAUG, we present case studies that show how we can optimize
overall performance and energy efficiency for up to 1.8-5$\times$ speedup over a baseline system,
without changing any part of the accelerator microarchitecture, as well as show how SMAUG can tune an SoC for a 
camera-powered deep learning pipeline.
\end{abstract}

\section{Introduction}
\label{sec:intro}
The tremendous popularity of deep learning (DL) in recent years has been fueled
by the increased capability of DL hardware and software systems. In particular,
for both performance and energy efficiency, dedicated hardware accelerators for
deep neural networks (DNNs) have received a phenomenal amount of interest
\cite{googletpu, eyeriss, judd2016stripes, chi2016prime,
liu2016cambricon,alwani2016fused,ding2017circnn}. Much of the focus on DNN
accelerator design has been on optimizing core datapaths and dataflows to
improve local reuse of data and reduce expensive data movement between the
processing elements, local storage, and DRAM, on a per-layer basis.  However, at the
end of the day, end-to-end performance is what truly matters, and additional
overheads must be considered, such as data layout transformations that shuffle
and reshape the data, the choice of accelerator interfacing with the SoC which
affects data movement efficiency, management of multiple independently
programmed accelerators, and contention for shared system resources like memory
bandwidth between different agents on the SoC.

\begin{figure}[t]
\centering
\includegraphics[width=0.5\textwidth, width=8cm]{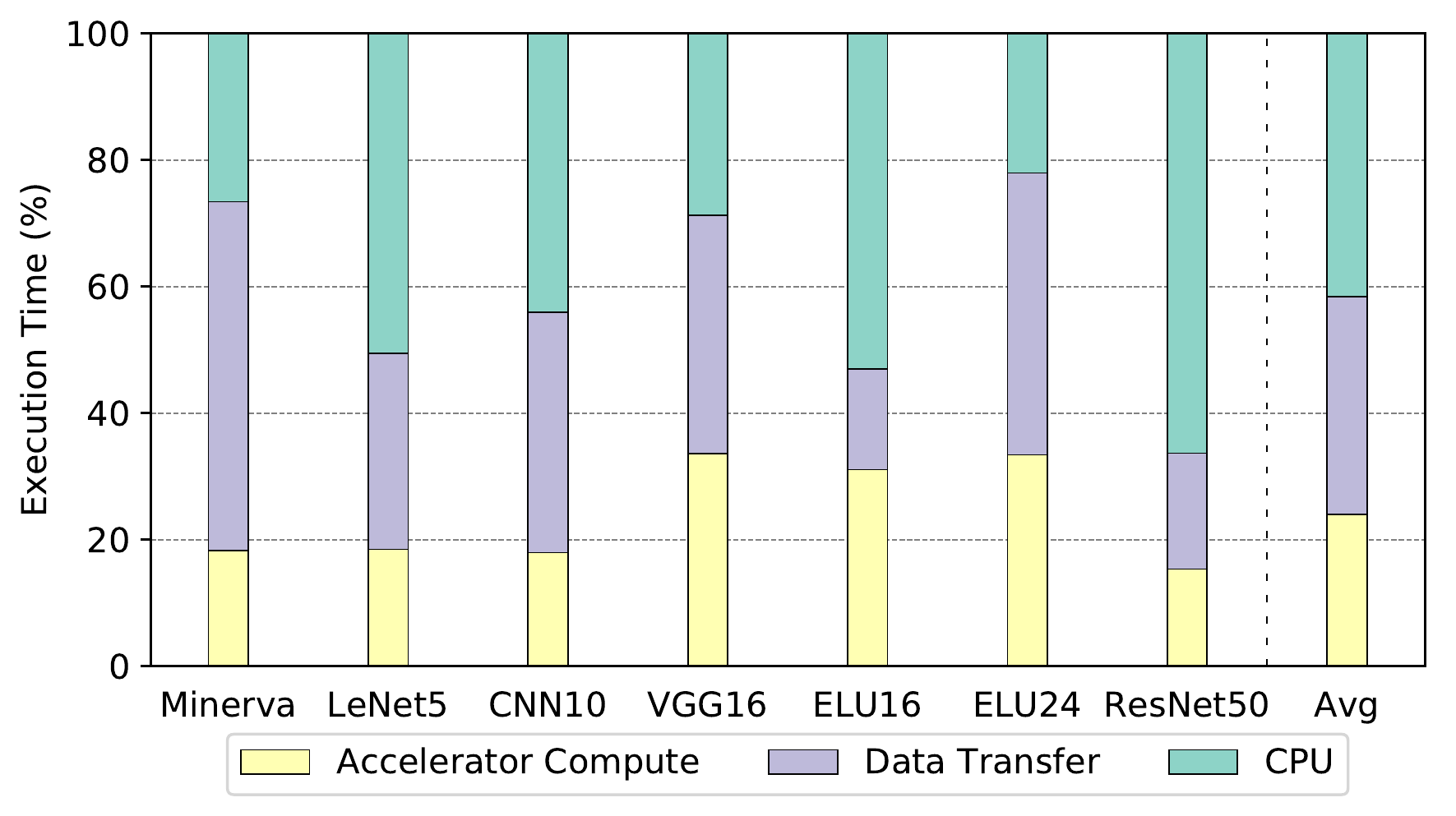}
\vspace{-0.3cm}
\caption{Overall inference latency on a range of DNNs is bottlenecked by data
  transfer and CPU processing because accelerators are already so well
  optimized.  Evaluated on SMAUG (see Section \ref{sec:method} for more details on the baseline SoC.)}
\vspace{-0.5cm}
\label{fig:baseline-cycles}
\end{figure}

As a motivating example of why overall performance is important, we profile
end-to-end single-batch inference latency on a range of image classification DNNs and
break down the overall time spent on accelerator compute, data transfer to/from
scratchpads, and CPU time spent in the software stack, on a system with one DNN
accelerator connected over DMA.  In this paper, we define the term
``accelerator'' to refer to any independently programmable hardware block
specialized for a few particular operations. Figure \ref{fig:baseline-cycles}
shows that out of the entire execution time, only $\sim$25\% on average is
spent waiting on accelerator compute, with the rest of the time taken up by
data transfer (34\%) and CPU processing (42\%), performing tasks like data
layout transformations, tiling, and more.  This is particularly the case on a
network like ResNet50 because of the many expensive tiling operations between
each of the 50 layers.  In some respects, this breakdown is not surprising
because the performance speedups offered by DNN accelerators can easily make
software the primary bottleneck of overall latency.  Nonetheless, this analysis
shows several opportunities for optimization that would not have been revealed
by a layer-by-layer analysis, which this paper will explore in more depth. The
impact of software stack time on overall performance has also been observed on
industry-grade deep learning models written in TensorFlow \cite{emmawang2019} and Caffe2 \cite{park2018deep}.

\begin{table}[t]
  \renewcommand\cellgape{\Gape[1.4pt]}
  \newcommand{\x}{\text{x}}
  \centering
  \footnotesize
  \resizebox{\columnwidth}{!}{
  \begin{tabular}{|c|c|c|c|c|c|} \hline
    & \thead{End-to-End \\ Evaluation} & \thead{Simulation \\ Support} & \thead{Easy Custom \\ Backend Integration} & \thead{Simulation \\ Speed} \\ \hhline{|=|=|=|=|=|=|}
    
    \makecell[c]{TensorFlow \cite{abadi2016tensorflow}, \\ PyTorch \cite{pytorch}, \\ Caffe/Caffe2 \cite{caffe}, \\ MXNet \cite{mxnet}} & \greencheck & \redcross & \redcross & N/A \\\hline
    
    \makecell[c]{DNNWeaver \cite{sharma2016dnnweaver}, \\ DNNBuilder \cite{zhang2018dnnbuilder}, \\ MAGNet \cite{venkatesanmagnet}} & \redcross & \makecell[c]{\greencheck \\ (accelerator TLM/RTL \\simulation only)} & \makecell[c]{\redcross \\ (requires RTL/HLS or\\detailed timing models)} & Varies \\\hline
    
    
    \makecell[c]{TVM/VTA \cite{TVM}} & \greencheck & \makecell[c]{\greencheck \\ (behavioral simulation for \\template accelerator)} & \makecell[c]{\redcross \\ (requires RTL/HLS)} & Fast \\\hline
    
    \makecell[c]{SCALE-Sim \cite{scale-sim}, \\ HSIM-DNN \cite{sun2019hsim}} & \redcross & \makecell[c]{\greencheck \\ (accelerator analytical \\model)} & \makecell[c]{\redcross \\ (backend specific)} & \makecell[c]{Fast} \\ \hline
    
    \makecell[c]{SMAUG} & \greencheck  & \makecell[c]{\greencheck\greencheck \\ (flexible accelerator and \\SoC modeling)} & \makecell[c]{\greencheck \\ (no RTL required)} & Fast \\\hline
    
  \end{tabular}
  }
  \caption{Deep Learning Research Infrastructure.}
  \label{tab:infrastructure}
  \vspace{-0.2cm}
\end{table}

Consequently, in order to holistically design DNN-centric SoCs,
we must be able to study end-to-end behavior in simulation, as simulation
is the usual methodology to evaluate early-stage/pre-RTL hardware designs.
However, as shown in Table \ref{tab:infrastructure}, there is no DNN
framework available that supports fast, early-stage design exploration in
simulation. Productivity-oriented frameworks like TensorFlow or PyTorch don't
support simulation at all, and the ones that do support end-to-end simulation
all require RTL/HLS for custom hardware accelerator integration, which is 
slow to write/generate and slow to simulate as well.


To address this gap in research infrastructure, we describe SMAUG: Simulating
Machine Learning Applications Using gem5-Aladdin. gem5-Aladdin is an SoC
simulator that supports modeling of complex heterogeneous SoCs
\cite{shao_micro2016}, and SMAUG is the first architecture-simulation friendly deep learning
framework.  We wrote SMAUG to be compatible with gem5-Aladdin because it is
built on the familiar gem5 simulator, supports flexible SoC, accelerator, and memory
topologies, and also does not require RTL for design space exploration of accelerators, all
of which greatly simplify the research and development process. SMAUG is
designed to enable DNN researchers to rapidly evaluate different accelerator
and SoC designs and perform hardware-software co-design, not to replace
existing frameworks.  SMAUG currently is targeted at
DNN inference, but we plan to incorporate support for training as well.

SMAUG's headline features include a Python API for easy network configuration,
support for a wide range of commonly used operators and network topologies,
various hardware accelerator implementations of core kernels with easy
plug-and-play of new custom operators, and a complete software stack that
manages operator tiling, multi-accelerator and multi-thread scheduling,
synchronization, and more.  In addition, SMAUG solves several key problems of
building architecture-simulation friendly deep learning frameworks:

\begin{enumerate}
    \item New accelerators are easy to implement in SMAUG's modular
      architecture. They can be implemented in just a few lines of code with
      Aladdin, or as a native gem5 object for more control over cycle-level
      timing.
    \item Running a complete forward pass through a DNN may require billions of
      operations, but for many core kernels, most of that work looks the same.
      SMAUG supports sampling of accelerator simulation through new easy-to-use
      Aladdin APIs, with error as low as 1\% for even the most aggressive
      sampling factors.
    \item Workarounds are provided for various simulator limitations to make up
      for the lack of complete OS feature support, such as a thread scheduler.
\end{enumerate}

To illustrate the capabilities of SMAUG and the kinds of insights that only
end-to-end DNN studies can provide, we present several case studies
demonstrating how to improve overall performance of various DNNs by 45-79\%
(1.8-$5\times$ speedup) over a baseline system without any changes to the
accelerator microarchitecture itself:

\begin{enumerate}
    \item Using different SoC-accelerator interfaces to achieve tighter
      coupling between the CPU and accelerators for 17-55\% overall speedup
      and up to 56\% energy wins.
    \item Using multiple independent accelerators to exploit tile-level
      parallelism in DNNs for 24-62\% overall speedup with eight
      accelerators over a single accelerator system.
    \item Using multithreading in the software stack to optimize data
      preparation time for up to 37\% overall speedup.
\end{enumerate}

Finally, we demonstrate how SMAUG can be integrated with a state-of-the-art
camera pipeline, implemented in Halide, to model even more complex applications
and identify opportunities for more efficient system design.

\section{SMAUG Framework}

\label{sec:smaug}

\begin{figure}[t]
\centering
\includegraphics[width=0.48\textwidth]{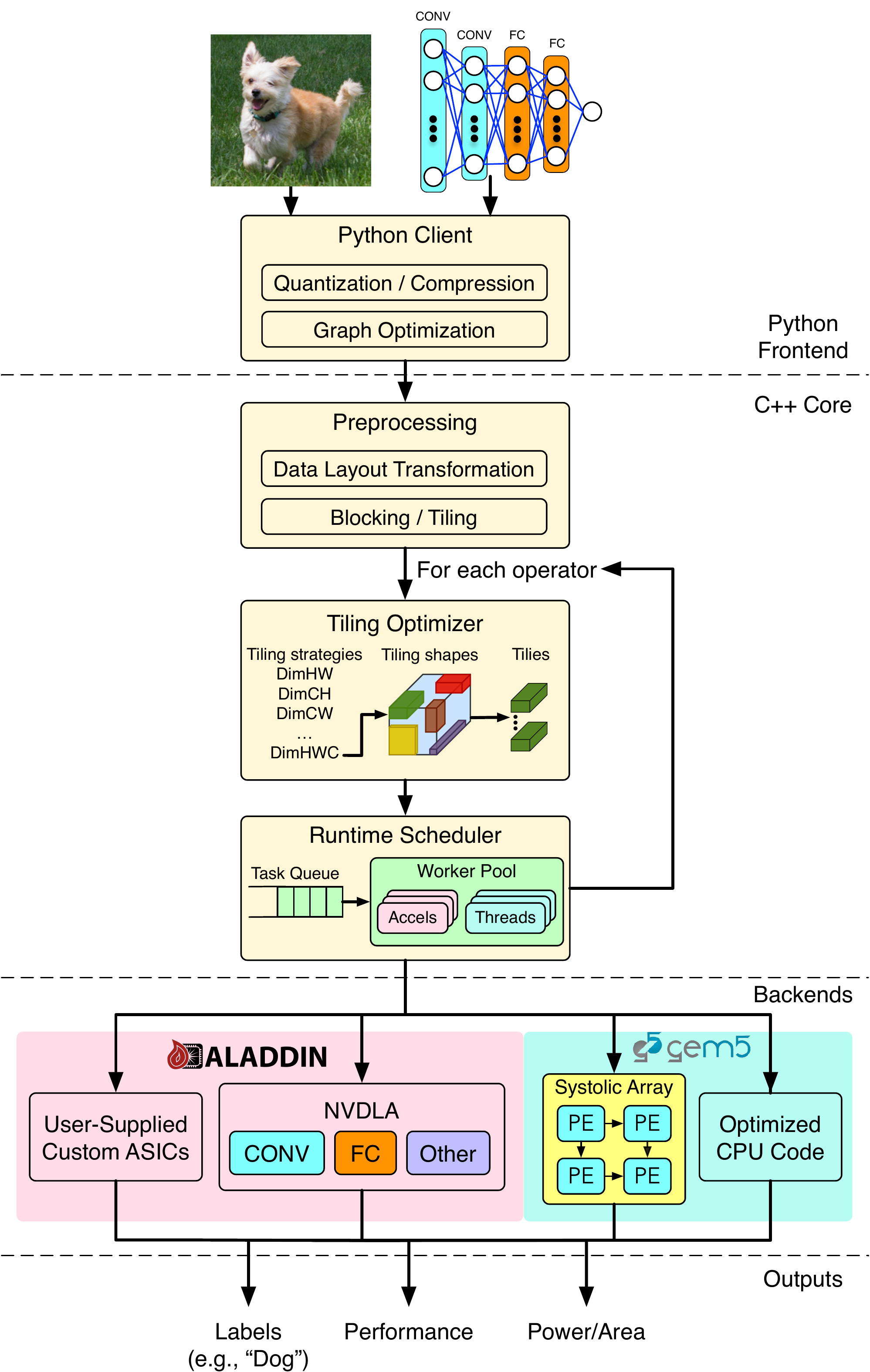}
\caption{Overview of SMAUG's execution flow.}
\vspace{-0.4cm}
\label{fig:smaug-overview}
\end{figure}

Figure \ref{fig:smaug-overview} illustrates the overall architecture and
execution flow of SMAUG. It is divided into three major components: a Python
frontend for network configuration, a C++ runtime to manage the execution flow,
and a backend consisting of a set of hardware-accelerated kernels. The
accelerated kernels can be modeled either using Aladdin or as a native gem5
simulation object, depending on the user's desired level of flexibility and
control.

Users begin by building a network using a declarative Python API, complete with all input
and weights data (which can optionally be taken from an existing trained
model). They also specify in the configuration which set of accelerated kernels
they want to use, the level of data quantization, and other metadata.  The
complete network specification is then converted into a dataflow graph and
serialized (along with network parameters).  This is a
one-time operation for each network, so it is done as a separate step.  The
serialized model is loaded into the C++ runtime, and SMAUG begins a set of
offline preprocessing steps. For example, certain read-only tensors (weights data) 
are pre-tiled (split into smaller contiguous tensors) during this preprocessing step
to reduce time on the critical path. This preprocessing can also be fast forwarded 
in simulation to save time. Next, SMAUG invokes a tiling optimizer to compute the
best available tiling shapes for each operation, so that the tiles utilize as much 
of the accelerator compute and memory resources as possible.  Finally, SMAUG 
dispatches each tile of work to the appropriate processing elements, waits for them 
to finish, gathers all the results, and prepares for the next operation. If
multiple independent PEs are available, SMAUG can schedule all of them at once.
Since the internal representation of the network is a graph, arbitrarily
complex networks can be defined and scheduled; the architecture is not limited
to linearly-stacked layers.  Multithreading support is available to parallelize
CPU operations when possible and better utilize available shared resources like
memory bandwidth.

As its name implies, SMAUG is compatible with the LLVM-based toolchains
required by the Aladdin accelerator simulator and the gem5 APIs it exposes
\cite{shao_micro2016}. We have made extensive improvements to these toolchains
to support compiling C++ binaries, tracing multi-threaded workloads, supporting
sampling, and more.  In the following sections, we will describe the frontend
Python API, core runtime, hardware backend modeling and simulation workarounds in more detail.


\subsection{Python API}
Many deep learning frameworks use Python APIs to build models, and we wanted to
follow in this same tradition to lessen the learning curve, rather than forcing
users to manually write configuration files or learn a new DSL. Figure
\ref{fig:python-api-example} shows how a residual unit might be built:

\begin{figure}[h]

\begin{Verbatim}[commandchars=\\\{\},fontsize=\scriptsize]
\PYG{k}{def} \PYG{n+nf}{create\PYGZus{}residual\PYGZus{}unit}\PYG{p}{():}
  \PYG{k}{with} \PYG{n}{Graph}\PYG{p}{(}\PYG{n}{name}\PYG{o}{=}\PYG{l+s+s2}{\PYGZdq{}residual\PYGZdq{}}\PYG{p}{,} \PYG{n}{backend}\PYG{o}{=}\PYG{l+s+s2}{\PYGZdq{}MyBackend\PYGZdq{}}\PYG{p}{)} \PYG{k}{as} \PYG{n}{g}\PYG{p}{:}
    \PYG{c+c1}{\PYGZsh{} Tensor initialization.}
    \PYG{n}{inputs} \PYG{o}{=} \PYG{n}{Tensor}\PYG{p}{(}\PYG{n}{np}\PYG{o}{.}\PYG{n}{random}\PYG{o}{.}\PYG{n}{rand}\PYG{p}{(}\PYG{l+m+mi}{1}\PYG{p}{,} \PYG{l+m+mi}{8}\PYG{p}{,} \PYG{l+m+mi}{32}\PYG{p}{,} \PYG{l+m+mi}{32}\PYG{p}{)))}
    \PYG{n}{filter0} \PYG{o}{=} \PYG{n}{Tensor}\PYG{p}{(}\PYG{n}{np}\PYG{o}{.}\PYG{n}{random}\PYG{o}{.}\PYG{n}{rand}\PYG{p}{(}\PYG{l+m+mi}{64}\PYG{p}{,} \PYG{l+m+mi}{3}\PYG{p}{,} \PYG{l+m+mi}{3}\PYG{p}{,} \PYG{l+m+mi}{3}\PYG{p}{)))}
    \PYG{n}{filter1} \PYG{o}{=} \PYG{n}{Tensor}\PYG{p}{(}\PYG{n}{np}\PYG{o}{.}\PYG{n}{random}\PYG{o}{.}\PYG{n}{rand}\PYG{p}{(}\PYG{l+m+mi}{8}\PYG{p}{,} \PYG{l+m+mi}{64}\PYG{p}{,} \PYG{l+m+mi}{3}\PYG{p}{,} \PYG{l+m+mi}{3}\PYG{p}{)))}
    \PYG{c+c1}{\PYGZsh{} If quantitization is desired:}
    \PYG{c+c1}{\PYGZsh{} filter0 = filter0.astype(np.float16)}
    \PYG{c+c1}{\PYGZsh{} Network topology:}
    \PYG{n}{act} \PYG{o}{=} \PYG{n}{input\PYGZus{}data}\PYG{p}{(}\PYG{l+s+s2}{\PYGZdq{}input\PYGZdq{}}\PYG{p}{,} \PYG{n}{input\PYGZus{}tensor}\PYG{p}{)}
    \PYG{n}{x} \PYG{o}{=} \PYG{n}{convolution}\PYG{p}{(}\PYG{l+s+s2}{\PYGZdq{}conv0\PYGZdq{}}\PYG{p}{,} \PYG{n}{act}\PYG{p}{,} \PYG{n}{filter0}\PYG{p}{,}
          \PYG{n}{stride}\PYG{o}{=}\PYG{p}{[}\PYG{l+m+mi}{1}\PYG{p}{,} \PYG{l+m+mi}{1}\PYG{p}{],} \PYG{n}{padding}\PYG{o}{=}\PYG{l+s+s2}{\PYGZdq{}same\PYGZdq{}}\PYG{p}{,} \PYG{n}{activation}\PYG{o}{=}\PYG{l+s+s2}{\PYGZdq{}relu\PYGZdq{}}\PYG{p}{)}
    \PYG{n}{x} \PYG{o}{=} \PYG{n}{convolution}\PYG{p}{(}\PYG{l+s+s2}{\PYGZdq{}conv1\PYGZdq{}}\PYG{p}{,} \PYG{n}{x}\PYG{p}{,} \PYG{n}{filter1}\PYG{p}{,}
          \PYG{n}{stride}\PYG{o}{=}\PYG{p}{[}\PYG{l+m+mi}{1}\PYG{p}{,} \PYG{l+m+mi}{1}\PYG{p}{],} \PYG{n}{padding}\PYG{o}{=}\PYG{l+s+s2}{\PYGZdq{}same\PYGZdq{}}\PYG{p}{)}
    \PYG{n}{out} \PYG{o}{=} \PYG{n}{add}\PYG{p}{(}\PYG{l+s+s2}{\PYGZdq{}add\PYGZdq{}}\PYG{p}{,} \PYG{n}{x}\PYG{p}{,} \PYG{n}{act}\PYG{p}{,} \PYG{n}{activation}\PYG{o}{=}\PYG{l+s+s2}{\PYGZdq{}relu\PYGZdq{}}\PYG{p}{)} \PYG{c+c1}{\PYGZsh{} residual}
    \PYG{k}{return} \PYG{n}{g}

\PYG{n}{graph} \PYG{o}{=} \PYG{n}{create\PYGZus{}residual\PYGZus{}unit}\PYG{p}{()}
\PYG{n}{graph}\PYG{o}{.}\PYG{n}{write\PYGZus{}graph}\PYG{p}{()} \PYG{c+c1}{\PYGZsh{} Graph serialization.}
\end{Verbatim}
\caption{Constructing a operations in SMAUG uses a familiar Python style.}
\label{fig:python-api-example}
\end{figure}

This small example demonstrates the simplicity and familiarity of
building networks in SMAUG.  They are specified in a deferred execution style,
and by using with-statement context managers, we can greatly reduce boilerplate
without adding global state. All input tensors must be constructed inside the
context before being used in an operator. Finally, the user can either supply
random data or existing trained parameters as well as the data type (e.g.
\texttt{float16} or \texttt{float32}). Certain optimizations like operator fusion
(e.g. convolution + element-wise operators) are applied automatically by the
framework. Finally, the user serializes the 
model specification and parameters; parameters are stored separately so that they
can be easily swapped.

\subsection{Tiling Optimizer} 
\label{sec:tiling-optimizer}
Due to the limited amount of local storage on an accelerator,
individual layers of DNNs often have too many weights and/or inputs to run
at once, thus requiring the operation to be tiled.  Whenever tiling is required,
redundant data movement is likely necessary, so identifying efficient tiling
schedules (also called ``loop nests'') that maximize data reuse and minimize
data movement between levels of the memory hierarchy is critical to
achieving high performance.  This has been studied extensively in the field;
however, the general problem of finding the optimal
solution is combinatorial in dimensionality and tiling factors, and
beyond the scope of this work.

In SMAUG, we circumvent a general solution by implementing a specialized tiling
optimizer for each dataflow implemented by an accelerator, because any
particular instantiation of an accelerator implements at most a few dataflows
which exploit particular parallelism patterns. As an example, consider the
dataflow of the Nvidia Deep Learning Accelerator (NVDLA), which we use in our
experiments (see Section \ref{sec:method} for details).  This dataflow,
described in Figure \ref{fig:nvdla-dataflow}, reduces partial products in the
channel dimension, so it benefits from a tiling schedule that maximizes the
channel dimension of the input and weight tiles. But such a schedule is not
suitable for an accelerator that computes 1D or 2D convolutions, like the
row-stationary dataflow \cite{eyeriss}.  By specializing the optimizer for a
specific dataflow, we restrict the search space to a narrower set of
possibilities that can be exhaustively explored. The final schedule is one that
maximizes both the utilization of the local scratchpads and the functional
units.

\begin{figure}[t]
\begin{Verbatim}[commandchars=\\\{\},fontsize=\scriptsize]
\PYG{n}{BUFFER} \PYG{n}{IN}\PYG{p}{[}\PYG{n}{IN\PYGZus{}R}\PYG{p}{][}\PYG{n}{IN\PYGZus{}C}\PYG{p}{][}\PYG{n}{IN\PYGZus{}H}\PYG{p}{];}
\PYG{n}{BUFFER} \PYG{n}{WGT}\PYG{p}{[}\PYG{n}{NUM\PYGZus{}PES}\PYG{p}{][}\PYG{n}{WGT\PYGZus{}R}\PYG{p}{][}\PYG{n}{WGT\PYGZus{}C}\PYG{p}{][}\PYG{n}{IN\PYGZus{}H}\PYG{p}{];}
\PYG{n}{BUFFER} \PYG{n}{OUT}\PYG{p}{[}\PYG{n}{NUM\PYGZus{}PES}\PYG{p}{][}\PYG{n}{OUT\PYGZus{}R}\PYG{p}{][}\PYG{n}{OUT\PYGZus{}C}\PYG{p}{];}
\PYG{n}{parallel} \PYG{n+nf}{for} \PYG{p}{(}\PYG{n}{pe} \PYG{o}{=} \PYG{l+m+mi}{0} \PYG{n}{to} \PYG{n}{NUM\PYGZus{}PES}\PYG{p}{)}
  \PYG{k}{for} \PYG{p}{(}\PYG{n}{kr} \PYG{o}{=} \PYG{l+m+mi}{0} \PYG{n}{to} \PYG{n}{WGT\PYGZus{}R} \PYG{o}{\PYGZhy{}} \PYG{l+m+mi}{1}\PYG{p}{)}
    \PYG{k}{for} \PYG{p}{(}\PYG{n}{kc} \PYG{o}{=} \PYG{l+m+mi}{0} \PYG{n}{to} \PYG{n}{WGT\PYGZus{}C} \PYG{o}{\PYGZhy{}} \PYG{l+m+mi}{1}\PYG{p}{)}
      \PYG{k}{for} \PYG{p}{(}\PYG{n}{cb} \PYG{o}{=} \PYG{l+m+mi}{0} \PYG{n}{to} \PYG{n}{IN\PYGZus{}H}\PYG{o}{/}\PYG{l+m+mi}{32} \PYG{o}{\PYGZhy{}} \PYG{l+m+mi}{1}\PYG{p}{)} \PYG{p}{\PYGZob{}}
        \PYG{c+c1}{// cb = channel block}
        \PYG{c+c1}{// Each PE has its own weight reg.}
        \PYG{n}{BUFFER} \PYG{n}{wgt\PYGZus{}reg}\PYG{p}{[}\PYG{l+m+mi}{0}\PYG{o}{:}\PYG{l+m+mi}{31}\PYG{p}{]} \PYG{o}{=} \PYG{n}{WGTS}\PYG{p}{[}\PYG{n}{pe}\PYG{p}{][}\PYG{n}{kr}\PYG{p}{][}\PYG{n}{kc}\PYG{p}{][}\PYG{n+nl}{cb}\PYG{p}{:}\PYG{n}{cb}\PYG{o}{+}\PYG{l+m+mi}{31}\PYG{p}{];}
        \PYG{c+c1}{// Now iterate over the input rows and cols.}
        \PYG{k}{for} \PYG{p}{(}\PYG{n}{r} \PYG{o}{=} \PYG{l+m+mi}{0} \PYG{n}{to} \PYG{n}{OUT\PYGZus{}R} \PYG{o}{\PYGZhy{}} \PYG{l+m+mi}{1}\PYG{p}{)}
          \PYG{k}{for} \PYG{p}{(}\PYG{n}{c} \PYG{o}{=} \PYG{l+m+mi}{0} \PYG{n}{to} \PYG{n}{OUT\PYGZus{}C} \PYG{o}{\PYGZhy{}} \PYG{l+m+mi}{1}\PYG{p}{)}
            \PYG{n}{parallel} \PYG{k}{for} \PYG{p}{(}\PYG{n}{h} \PYG{o}{=} \PYG{l+m+mi}{0} \PYG{n}{to} \PYG{l+m+mi}{31}\PYG{p}{)} \PYG{p}{\PYGZob{}}
              \PYG{c+c1}{// 32\PYGZhy{}way spatial reduction in channel}
              \PYG{c+c1}{// dimension.}
              \PYG{n}{OUT}\PYG{p}{[}\PYG{n}{r}\PYG{p}{][}\PYG{n}{c}\PYG{p}{][}\PYG{n}{pe}\PYG{p}{]} \PYG{o}{+=} \PYG{n}{IN}\PYG{p}{[}\PYG{n}{r}\PYG{o}{+}\PYG{n}{kr}\PYG{p}{][}\PYG{n}{c}\PYG{o}{+}\PYG{n}{kc}\PYG{p}{][}\PYG{n}{cb}\PYG{o}{*}\PYG{l+m+mi}{32}\PYG{o}{+}\PYG{n}{h}\PYG{p}{]} \PYG{o}{*}
                              \PYG{n}{wgt\PYGZus{}reg}\PYG{p}{[}\PYG{n}{h}\PYG{p}{];}
            \PYG{p}{\PYGZcb{}}
      \PYG{p}{\PYGZcb{}}
\end{Verbatim}
\caption{Dataflow implemented by NVDLA. Apart from syntax, this is
nearly the actual C code in SMAUG.}
\vspace{-0.4cm}
\label{fig:nvdla-dataflow}
\end{figure}

The first major step of the tiling optimizer is to identify a \emph{tiling
strategy}, i.e. the best dimensions along which to tile the input and output
tensors.  If an input tensor is of shape $N*H*W*C$, there are four possible
dimensions to tile the tensor, and the best choice depends on two
factors. The first is the accelerator: the dataflow it implements and the
minimum number of elements along the dimensions that maximize usage of the
functional units. For example, if the accelerator reduces partial products
along the channel dimension, and it implements a 32-way reduction unit, each
tile should have a multiple of 32 channels while also fitting inside the
accelerator's scratchpads.  The second factor is data layout of the tensor,
which determines the amount of work required to shuffle and reshape the
original input tensor into these tiles. This is a consideration that only arises
when evaluating end-to-end performance.

\begin{figure}[t]
\centering
\includegraphics[width=7.8cm]{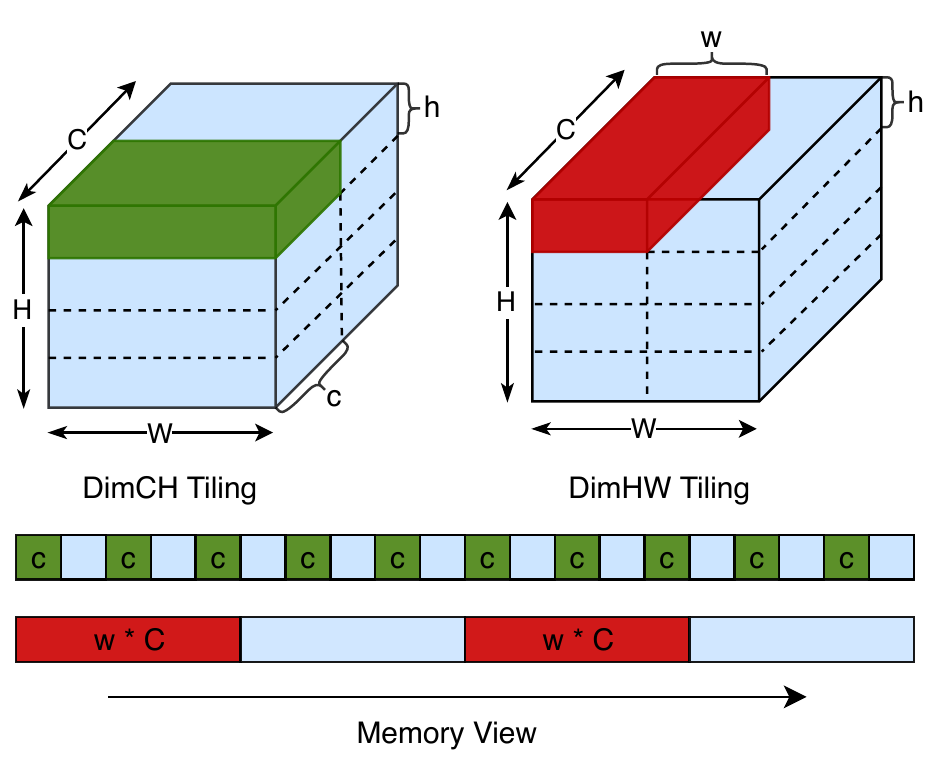}
\vspace{-0.2cm}
\caption{An NHWC tensor can be tiled using different tiling strategies.}
\label{fig:tiling}
\end{figure}

\begin{figure}[t]
\centering
\includegraphics[width=0.38\textwidth]{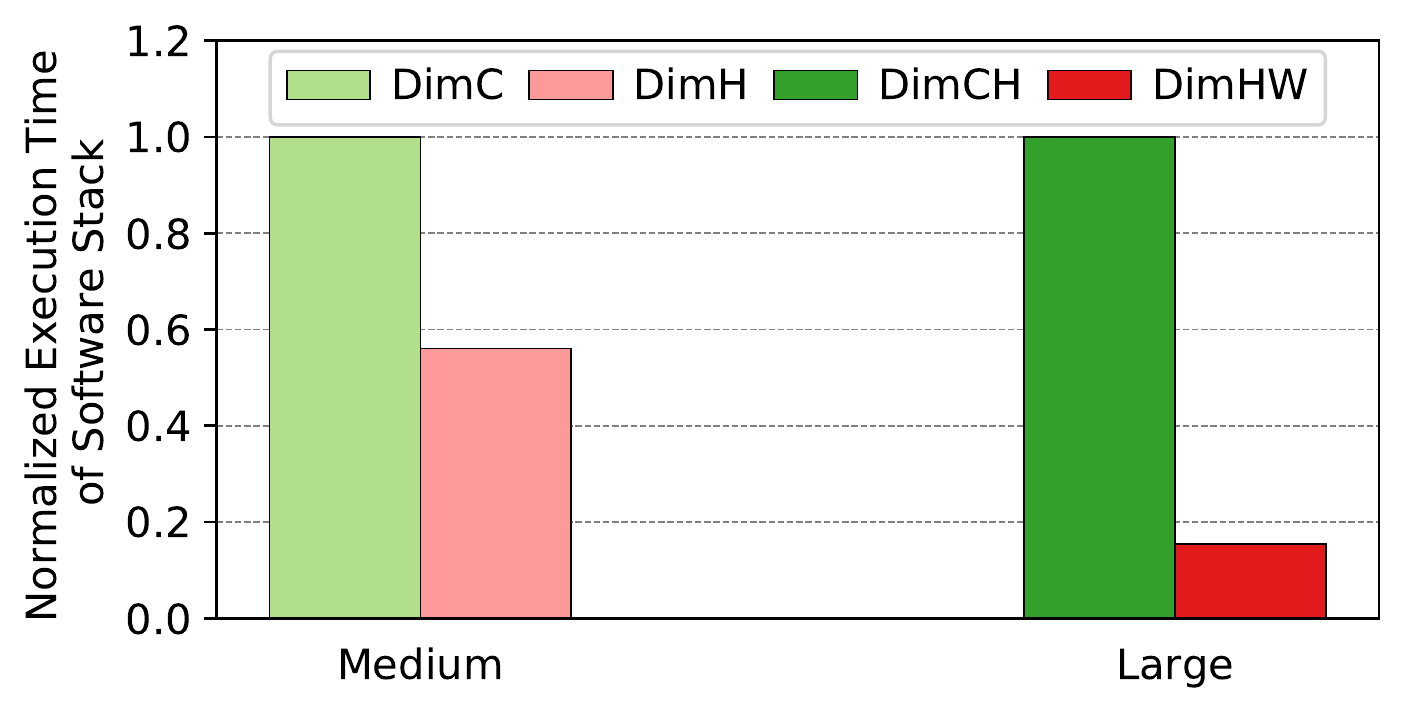}
\caption{Different tiling strategies for a medium (1x16x16x128) and large
  NHWC tensor (1x64x64x512) exhibit very different transformation costs.}
\label{fig:tiling-tradeoff}
\vspace{-0.2cm}
\end{figure}

As an example, Figure \ref{fig:tiling} shows how one NHWC tensor, when tiled in
two different ways, produces two very different memcpy patterns.  We describe a
tiling strategy with the notation \texttt{DimXYZ}, where X, Y, and Z are the
tiled dimensions. For this tensor, channels are the innermost dimension, so
it is the most expensive to tile. To quantify this difference, we show in
Figure \ref{fig:tiling-tradeoff} how long it takes to tile two different
tensors two different ways for a max tile size of 16,384 elements.
The medium-sized tensor (1x16x16x128) can be
tiled channel-wise (1x16x16x64) or row-wise (1x8x16x128), but row-wise is
$1.78\times$ faster in software because it only requires two large memcpys of
8x16x128 = 16K contiguous elements, whereas for channel-wise tiling requires
512 memcpys of 64 elements. This effect is even more pronounced on the larger
layer, where we can use either DimCH (1x32x64x8) or DimHW (1x1x32x512). DimHW
tiling is $6.5\times$ faster to complete because it only requires 128 memcpys
of 16K elements to completely tile the input, compared to 262K memcpys of 8
elements. The effect of a different tiling strategy on the overall operation is
harder to predict. For element-wise operations, tiling strategy has next to no
effect; for operations whose performance depends on exploiting data reuse,
changing tiling shape may impact overall runtime. This is one of the new
tradeoffs SMAUG enables researchers to explore.


The second major step of the tiling optimizer is to compute the best tile shapes
given the tiling strategy and max tile size. Depending on the chosen strategy and
operator parameters, this process is surprisingly complex and can encounter an
incredible number of edge cases, all of which must be handled for correctness
and efficiency.  Considerations include halo regions around the entire tensor
due to zero-padding, overlapping regions around each tile, interactions with > 1 stride sizes,
and more. In most cases, input tile shapes are not uniform and thus produce differently sized output
tiles.

\subsection{Runtime Scheduler}

Once the tiling shapes have been generated for each operator's tensors, the
scheduler prepares the tensor for computation by splitting it into the specified
tile shapes. This has to wait until runtime, when the actual input data from the previous
operator is ready. Then, the scheduler dispatches each tile to the appropriate compute element.
If there are multiple compute elements and tile-level parallelism exists, SMAUG can dispatch
independent work to multiple compute elements at once.
To help distribute work across multiple accelerators, SMAUG implements an accelerator worker pool and a command
queue per accelerator. Tasks are pushed onto the command queue for the next
available accelerator in the pool.
Any operators that are not
supported in the backend hardware accelerators are executed on the CPU
instead. As the work completes, the scheduler gathers the tiled output data back
into one contiguous tensor. Since the gem5-Aladdin API
exposes an accelerator like a thread, managing multiple accelerators concurrently
running is as straightforward as starting and joining on threads.
SMAUG also supports dividing CPU work across multiple threads, but since
gem5's syscall-emulation mode does not have a thread scheduler, SMAUG
implements a thread pool with round-robin scheduling of tasks from a work queue.

\subsection{Backends}
The backends run the convolutions/inner products/etc. required by the model. In SMAUG,
we provide a complete set of hardware accelerated kernels for all the included operators.
These models can be written using Aladdin or as a native gem5 object, depending on
the user's desired level of flexibility and control. We provide examples of both, most
notably a convolution engine inspired by NVDLA, written with Aladdin, and a configurable
systolic array, written as a native cycle-level gem5 object.

The NVDLA-inspired convolution engine consists of eight PEs, each with a 32-way multiply-accumulate (MACC) array that operates
on a different output feature map. The dataflow is described in Figure \ref{fig:nvdla-dataflow}.
Inputs and weights are 16-bit fixed point, while outputs are accumulated in 32-bit fixed point and reduced to 16-bit before
being written to the scratchpad.
In the emerging vernacular used to describe DNN dataflows, this dataflow is L0 weight-stationary
(weights are reused every cycle at the register level within a MACC array), and L1 input/output stationary (for every weight,
inputs are re-read and outputs are accumulated in-place in the SRAMs).
It is backed by three SRAMs, one each for inputs, weights, and outputs. We only model the core datapath and dataflow of NVDLA,
not other features like its convolution buffer.

The systolic array's dataflow is output stationary: inputs stream through from the left,
while weights stream from the top. There
are three scratchpads, accessed from fetch and commit units, to supply the PEs with data.
The dataflow was inspired by SCALE-Sim \cite{scale-sim}, but SCALE-Sim is
primarily an analytical model that can generate SRAM and memory traces to feed to other
tools like DRAMSim, whereas our model is entirely execution-driven and produces live memory
traffic that affects (and is affected by) the rest of the system.

One of the key design features of SMAUG is how easy it is to implement a
new HW accelerator model and integrate into the framework, particularly if
the user chooses to use Aladdin for modeling.
For example, apart from syntax and variable declarations, Figure \ref{fig:nvdla-dataflow} is very
similar to the code that models the convolution engine. In fact, merely 5\%
of the code is used to model all the hardware blocks with Aladdin.
If users do choose to write cycle-level timing models using native gem5 APIs,
then more code is needed (the systolic array model accounts for $\sim$10\% of 
the SMAUG codebase). The remaining 85\% is
devoted to computing tiling schedules, memory management, data movement,
cache coherency, and task scheduling. Therefore, SMAUG eases the development
of not only new hardware models, but also studies of end-to-end system interactions,
enabling researchers to spend their time on the topics that interest them the most.


\subsection{Working with Simulator Limitations}
In order to simulate end-to-end networks in user-level simulators (like gem5 syscall-emulation mode),
there are three constraints that must be addressed: reducing simulation time with sampling,
handling incomplete system call emulation, and supporting multi-threading without
a thread scheduler (typically implemented in the kernel).

\subsubsection{Sampling Support}
Modern DNNs are very deep and compute intensive, often requiring billions of
operations, and because Aladdin is a trace based simulator, it may be
infeasible to store and simulate a complete forward pass. However, since DNN computation is so regular, a sampling approach works well.

\begin{figure}[t]
\begin{Verbatim}[commandchars=\\\{\},fontsize=\scriptsize]
\PYG{k+kt}{int} \PYG{n+nf}{reduction}\PYG{p}{(}\PYG{k+kt}{int}\PYG{o}{*} \PYG{n}{a}\PYG{p}{,} \PYG{k+kt}{int} \PYG{n}{size}\PYG{p}{,} \PYG{k+kt}{int} \PYG{n}{sample}\PYG{p}{)} \PYG{p}{\PYGZob{}}
    \PYG{c+c1}{// Generally avoid sampling loops containing data}
    \PYG{c+c1}{// transfer operations to avoid changing the memory}
    \PYG{c+c1}{// footprint of the application.}
    \PYG{n}{dmaLoad}\PYG{p}{(}\PYG{n}{a}\PYG{p}{,} \PYG{n}{size} \PYG{o}{*} \PYG{k}{sizeof}\PYG{p}{(}\PYG{k+kt}{int}\PYG{p}{));}
    \PYG{k+kt}{int} \PYG{n}{result} \PYG{o}{=} \PYG{l+m+mi}{0}\PYG{p}{;}
    \PYG{n}{setSamplingFactor}\PYG{p}{(}\PYG{l+s}{\PYGZdq{}loop\PYGZdq{}}\PYG{p}{,} \PYG{p}{(}\PYG{k+kt}{float}\PYG{p}{)}\PYG{n}{size} \PYG{o}{/} \PYG{n}{sample}\PYG{p}{);}
    \PYG{n+nl}{loop}\PYG{p}{:}
    \PYG{c+c1}{// Run only `sample` iterations of this loop; the result}
    \PYG{c+c1}{// might be wrong, but that\PYGZsq{}s expected for sampling.}
    \PYG{k}{for} \PYG{p}{(}\PYG{k+kt}{int} \PYG{n}{i} \PYG{o}{=} \PYG{l+m+mi}{0}\PYG{p}{;} \PYG{n}{i} \PYG{o}{\PYGZlt{}} \PYG{n}{sample}\PYG{p}{;} \PYG{n}{i}\PYG{o}{++}\PYG{p}{)}
        \PYG{n}{result} \PYG{o}{+=} \PYG{n}{a}\PYG{p}{[}\PYG{n}{i}\PYG{p}{];}
    \PYG{k}{return} \PYG{n}{result}\PYG{p}{;}
\PYG{p}{\PYGZcb{}}
\end{Verbatim}
\caption{An example of specifying sampling factors on loops in Aladdin.}
\label{fig:sampling}
\end{figure}

\begin{figure}[t]
\centering
\includegraphics[width=8cm]{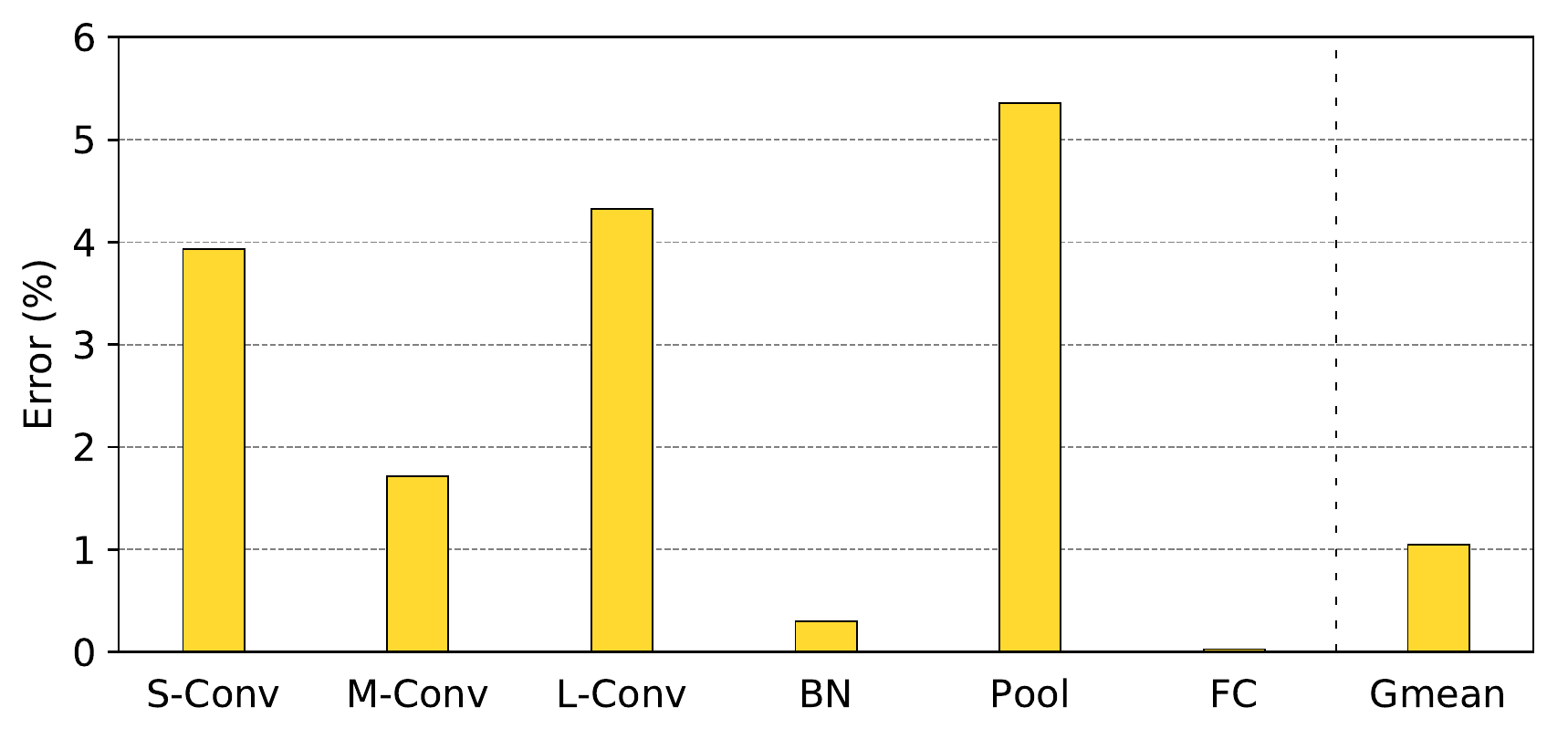}
\caption{Sampling performance validation. S-Conv uses 16 1x1x8 kernels; M-Conv
  uses 64 2x2x16 kernels; L-Conv uses 256 3x3x64 kernels.}
\label{fig:sampling-error}
\end{figure}

We extended Aladdin to support sampling at the per-loop level with a new API
called \texttt{setSamplingFactor}, in which the user specifies
how many iterations of the loop to trace and simulate.  Figure
\ref{fig:sampling} demonstrates the API at work. During Aladdin's graph
optimization process, we build a loop tree that captures the hierarchy of loop
iterations. When the simulation is over, Aladdin examines each node in the loop
tree, unsamples the latency of simulated iterations, and propagates the sampled
execution time up the tree. After every loop is unsampled, Aladdin
produces a final overall cycles estimate. This API supports pipelined loops as
well, although at least two loop iterations are required to determine the
pipeline latency. As a result, we can simulate a forward pass of ResNet50 in
just 5 hours.  In our experiments, we only sample loops containing only computation, 
not loops containing large data transfers, so that the memory footprint of the sampled network is
unchanged.  We validated our sampling technique
for a range of operators and input shapes, all at the highest sampling factors
(so that the sampled loops only run one or two iterations). Figure
\ref{fig:sampling-error} shows that sampled execution has less than 6\% error
across different kernel types, with an average of just 1\%. Finally, note that
sampled simulation will obviously produce incorrect functional results, since not
all the code is being executed, so this is only suitable for loops whose control
flow is not data dependent.

\subsubsection{System call emulation}
User-space simulators often do not implement the full range of OS features that
we come to take for granted. For example, the
\texttt{mmap} syscall can map the contents of a file into memory
(among other use cases), so it can be manipulated directly via loads and stores
rather than through the IO subsystem, but in gem5 syscall-emulation mode, stores to mmapped
memory are not synchronized to the backing file.  SMAUG was written to work
within all of these limitations; it compiles into a single C++ binary, never
forks other processes, and interacts minimally with the OS, essentially only
requiring the ability to access a filesystem, call \texttt{printf}, and start
new threads that never exit.

\subsubsection{Multithreading support}
\label{sec:multithreading}
It is common for user-level simulators to not implement pre-emptive thread
schedulers, as thread scheduling is a kernel task. gem5 syscall emulation mode
has limited support for multi-threading, with limitations on how thread
contexts can be reused after a thread exits. To enable multi-threading in
SMAUG, we implemented a custom thread pool and expose an API to dispatch work
to it.  Each task is executed to completion before yielding the CPU.
Furthermore, to prevent idle threads from spinning endlessly in simulation and
generating useless work that slows down the simulation, we use gem5 hooks to
quiesce CPUs while they're waiting for work and wake them only when we assign
them tasks.

\section{Methodology}
\label{sec:method}

Now that we have described SMAUG, we will demonstrate how it can be used to provide insights into accelerated DNN performance that per-layer studies would not be able to show. First, we discuss our evaluation methodology.

\subsection{Baseline System}

\begin{figure}[t]
\centering
\includegraphics[width=0.5\textwidth, width=8cm]{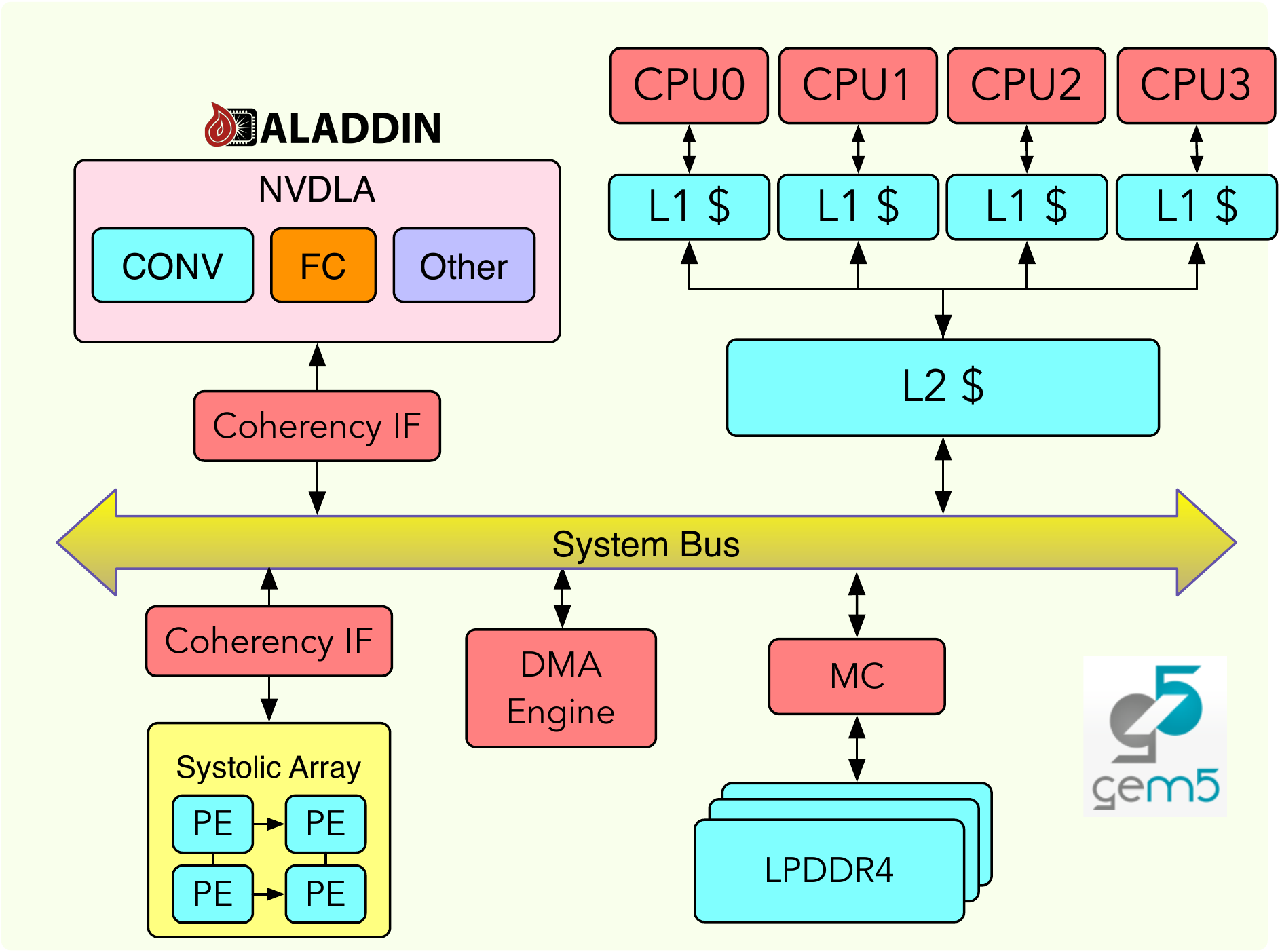}
\caption{The SoC platform used in this paper's experiments.}
\label{fig:soc-overview}
\end{figure}

\begin{table}[h]
\centering
\scriptsize
\begin{tabular}{|l|l|}
\cline{1-2}
Component & Parameters \\\hhline{|=|=|}
CPU Core   & \makecell[l]{8 Out-of-order X86 cores @2.5GHz \\ 8-\si{\micro}op issue width, 192-entry ROB} \\\hline
L1 Cache   & \makecell[l]{64KB i-cache \& d-cache, 4-way associative, 32B cacheline, \\ LRU, 2-cycle access latency} \\ \hline
L2 Cache   & \makecell[l]{2MB, 16-way, LRU, MESI coherence, 20-cycle access latency} \\ \hline
DRAM       & \makecell[l]{LP-DDR4, @1600MHz, 4GB, 4 channels, 25.6GB/s} \\ \hline
Accels     & \makecell[l]{NVDLA conv engine and others, systolic array (8x8 PEs), 1GHz\\ All scratchpads are 32KB each} \\ \hline
\end{tabular}
\vspace{-0.2cm}
\caption{SoC microarchitectural parameters.}
\vspace{-0.2cm}
\label{tab:parameters}
\end{table}

Figure \ref{fig:soc-overview} shows the baseline SoC used in this paper, with microarchitectural parameters listed in Table 
\ref{tab:parameters}. In gem5-Aladdin, we use syscall-emulation mode with Ruby to model a MESI coherency protocol. The CPU 
communicates with the accelerator either via the \texttt{ioctl} system call or via shared memory. The baseline SoC transfers data over DMA and runs
a single-threaded software stack. In Section 
\ref{sec:case-study-dnn} (also Figure \ref{fig:baseline-cycles}), we run the convolutions and inner products on the NVDLA-inspired accelerator; in Section 
\ref{sec:case-study-camera}, we use the systolic array instead for diversity.

\subsection{Workloads}
\begin{table}[t]
  \renewcommand\cellgape{\Gape[1.4pt]}
  \newcommand{\x}{\text{x}}
  \centering
  \footnotesize
  \resizebox{\columnwidth}{!}{
  \begin{tabular}{|c|c|l|c|c|} \hline
    \makecell[c]{\textbf{Name}}  & \textbf{Dataset} & \textbf{Network Topology} & \makecell[c]{\textbf{Parameters}} & \textbf{Accuracy} \\ \hhline{|=|=|=|=|=|}
    
    Minerva \cite{minerva} & \makecell[c]{MNIST \\ (28x28x1)} & 4 FC [784, 256, 256, 10]. & 665KB & 97\% \\\hline
    
    LeNet5 \cite{lecun_ieee98} & \makecell[c]{MNIST \\ (28x28x1)} & \makecell[l]{5 layer CNN (3x3) \\ 2 CONV [32, 32], POOL, \\ FC [128, 10].} & \makecell[c]{1.2MB} & 98\% \\\hline
    
    CNN10 & \makecell[c]{CIFAR-10 \\ (32x32x3)} & \makecell[l]{10 layer CNN (3x3) \\ 4 CONV [32, 32, 64, 64], 2 BN, 2 POOL, \\ 2 FC [512, 10].} & \makecell[c]{4.2MB} & 85\% \\\hline
    
    VGG16 \cite{cifar10_vgg} & \makecell[c]{CIFAR-10 \\ (32x32x3)} & \makecell[l]{16 layer CNN (3x3). \\ 2 CONV [64, 128], POOL, \\ 2 CONV [128, 128], POOL, \\ 3 CONV [256, 256, 256], POOL, \\ 3 CONV [512, 512, 512], POOL, \\ 2 FC: [512, 10]. } & \makecell[c]{17.4MB} & 90\%  \\\hline
    
    ELU16 \cite{clevert2013elu} & \makecell[c]{CIFAR-100 \\ (32x32x3)} & \makecell[l]{16 layer CNN. \\ 1 CONV [192], POOL, \\ 2 CONV [192, 240], POOL, \\ 2 CONV [240, 260], POOL, \\ 2 CONV [260, 280], POOL, \\ 2 CONV [280, 300], POOL, \\ 2 CONV [300, 100]. \\ Mostly 1x1 \& 2x2 CONV.} & 3.3MB & 71.25\%  \\\hline
          
    ELU24 \cite{clevert2013elu} & \makecell[c]{CIFAR-100 \\ (32x32x3)} & \makecell[l]{24 layer CNN. \\ CONV [384], POOL, \\  4 CONV [384, 384, 640, 640], POOL, \\  4 CONV [640, 768, 768, 768], POOL, \\  3 CONV [768, 896, 896], POOL, \\ 3 CONV [896, 1024, 1024], POOL, \\ 4 CONV [1024, 1152, 1152, 100]. \\ Mostly 1x1 \& 2x2 CONV. } & 75MB & 77.72\% \\ \hline
                        
    ResNet50 \cite{resnet} & \makecell[c]{ImageNet \\ (224x224x3)} & \makecell[l]{50 layer CNN. \\
    1 CONV [64], \\
    3 stacks of 3 CONV [64, 64, 256], \\
    4 stacks of 3 CONV [128, 128, 512], \\
    6 stacks of 3 CONV [256, 256, 1024], \\
    3 stacks of 3 CONV [512, 512, 2048], \\
    1 FC [1000]. \\ 1x1 \& 3x3 CONV.} & 237MB & 76.46\% \\ \hline
  \end{tabular}
  }
  \caption{Datasets and networks used in this paper. All parameters are stored as 16 bit
  fixed-point.}
  \label{tab:networks}
  \vspace{-0.2cm}
\end{table}

With the flexible Python client and the complete SW/HW stack in SMAUG, we are able to evaluate a variety of DNN workloads. Here we investigate four image classification tasks: MNIST, CIFAR10, CIFAR100 and ImageNet. For the first three datasets, we select two different networks each. For ImageNet, we use ResNet50 \cite{resnet} (included in the emerging MLPerf Inference Benchmark \cite{reddi2019mlperf}). Table \ref{tab:networks} summarizes the networks used. The goal was to cover a diverse set of network topologies that still map well to the accelerator's dataflow, which is optimized for convolution shapes deep in input/output feature maps.

\subsection{Simulation Time}

\begin{figure}[t]
\centering
\includegraphics[width=8cm]{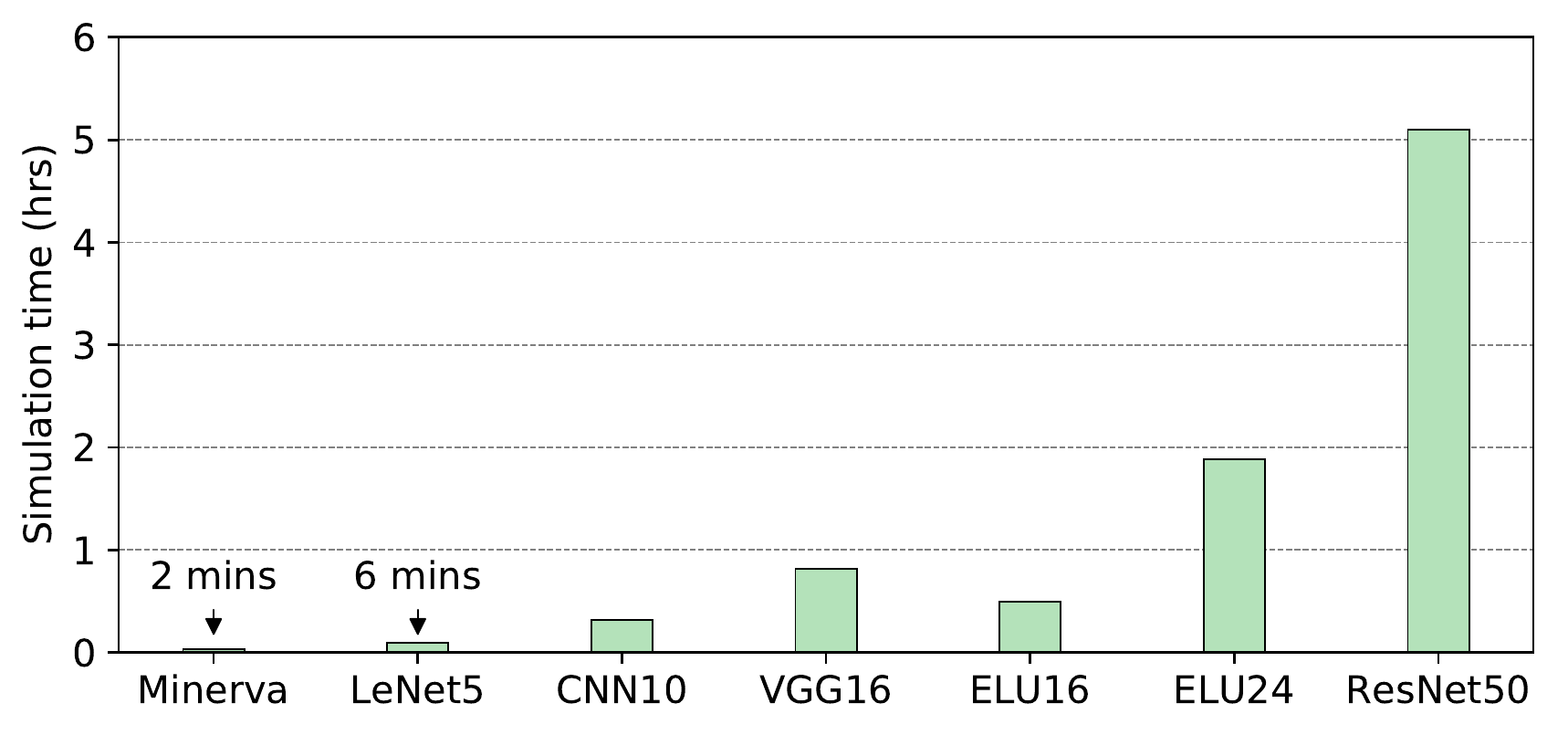}
\vspace{-0.2cm}
\caption{With sampling, even large networks can be simulated in hours.}
\vspace{-0.4cm}
\label{fig:simulation-time}
\end{figure}

Figure \ref{fig:simulation-time} shows the simulation time for running the workloads using the NVDLA backend in SMAUG on an Intel Xeon E5-2697 host (@2.6GHz). For most of the networks, SMAUG simulations finish within 2 hours. The smaller MNIST workloads take less than 10 minutes, and with sampling, even the large ResNet50 network finishes in $\sim$5 hours.

\subsection{Power and Area Modeling}
To obtain power and area estimates, we take a multi-pronged approach. We characterize various 16-bit functional units for power and area in a commercial 16nm FinFET process and plug them directly into Aladdin. To model accelerator local scratchpads, we build and characterize a variety of SRAM blocks through a commercial memory compiler in the same technology node. LLC power estimates are obtained from CACTI 7 \cite{jouppi2015cacti}, and DRAM power is modeled by DRAMPower \cite{drampower}, with timing and power parameters taken from a commercial LP-DDR4 product datasheet \cite{micron_lpddr4}.
\section{Optimizing End-to-End Performance of DNN workloads}
\label{sec:case-study-dnn}

SMAUG enables a wide range of architecture simulation tasks, from diverse DNN
topologies to accelerator implementations, from the SoC integration of
accelerators to evaluation of multi-accelerator systems. To illustrate the
insights that SMAUG can bring to DL hardware architects, in this section we demonstrate
several ways to improve end-to-end DNN performance on an SoC, all without
changing the underlying accelerator themselves.

In these case studies, the baseline system uses one NVDLA accelerator with the
DMA interface, running on a single-threaded software stack. Figure
\ref{fig:baseline-cycles} has shown that
not only the accelerator compute, but also data movement and CPU processing
spent on ``between-the-layer'' work are crucial to end-to-end DNN performance.
Therefore, in the rest of this section, we present three case studies that
attack all these components of performance. First, we optimize data transfers
by using a one-way coherent interface between the accelerator and SoC instead
of DMA. Second, we explore multi-accelerator systems to exploit tile-level
parallelism for greater compute and data-transfer throughput. Third, we
optimize tiling transformations in software to reduce CPU processing time. As a
whole, these optimizations speed up overall inference latency by 1.8-$5\times$.

\subsection{Improving Data Transfer: Coherent Accelerator-SoC Interfaces}

\begin{figure}[t]
\centering
\subfloat[Execution time.]{
	\label{fig:soc-interface-cycles}
	\includegraphics[width=0.5\textwidth,width=8cm]{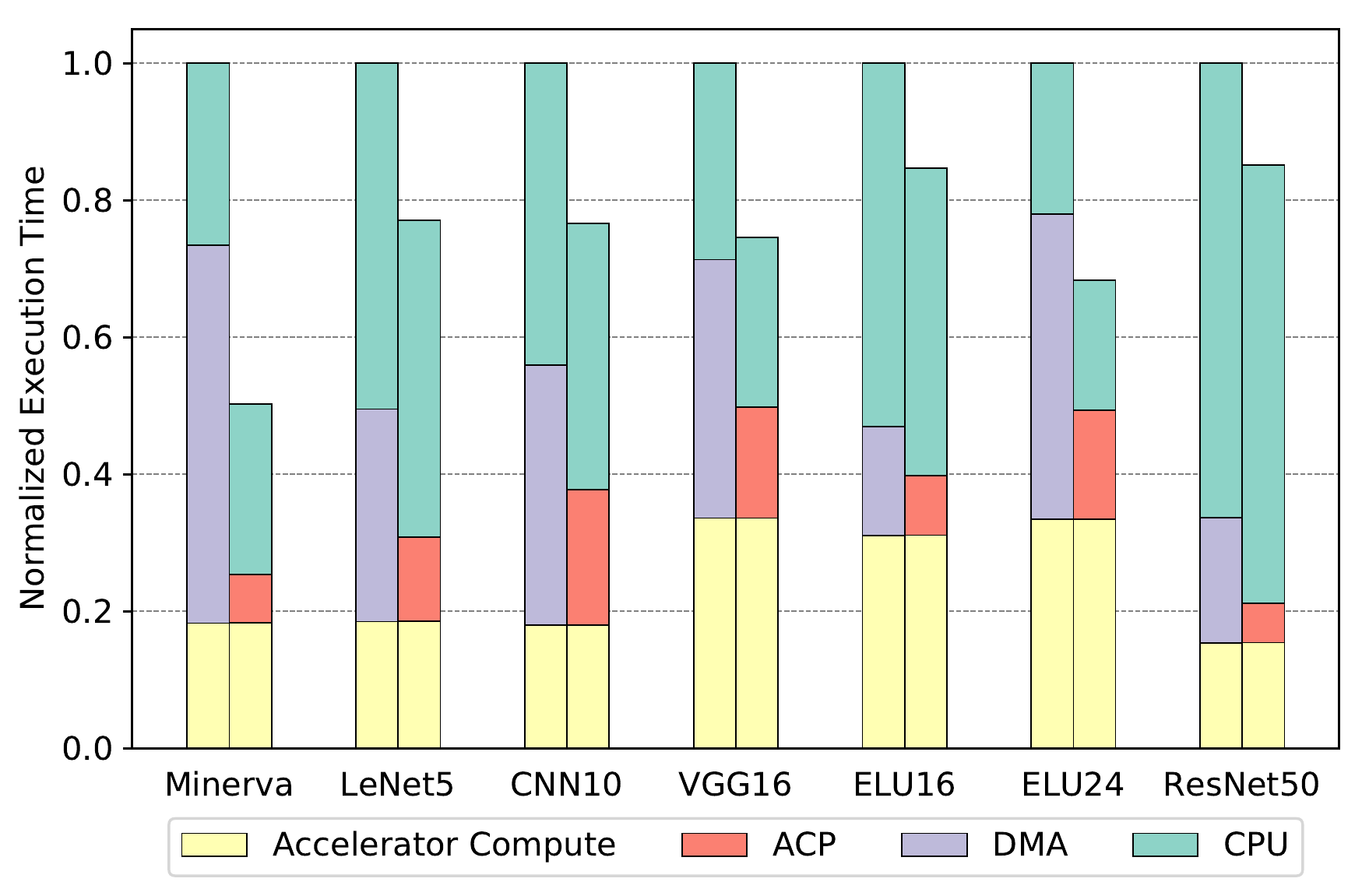}}
\hfill
\subfloat[Energy.]{
	\label{fig:soc-interface-energy}
	\includegraphics[width=0.5\textwidth,width=8cm]{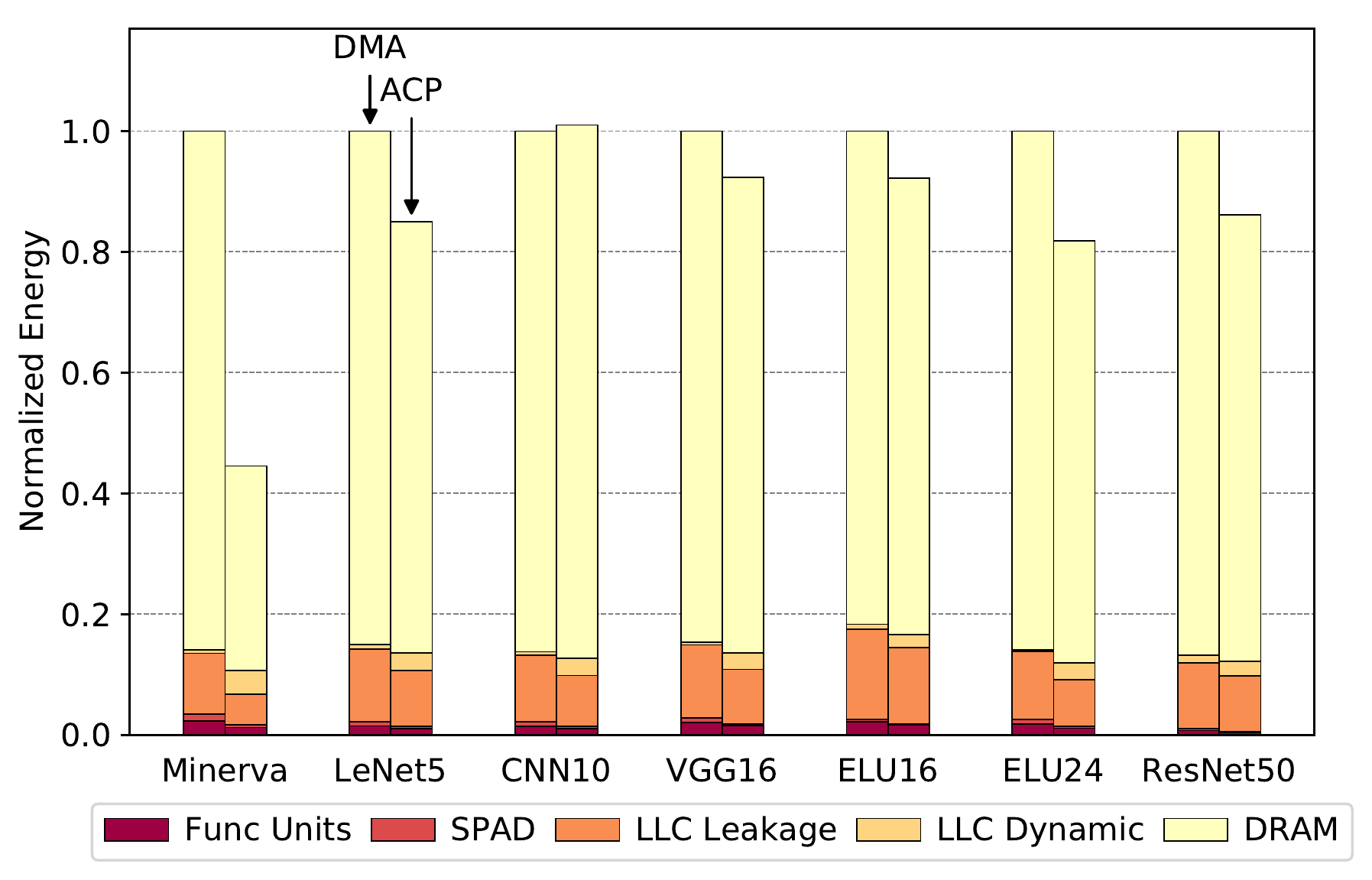}}
\caption{Performance and energy of the ACP interface, compared to DMA.}
\label{fig:soc-interface}
\vspace{-0.4cm}
\end{figure}

Fixed-function accelerators typically use private scratchpads for local storage
and communicate with the memory system of the SoC through a software-managed
DMA interface. DMA is the simplest approach to sharing data from a hardware
point-of-view, but it requires software to be responsible for explicitly
flushing and invalidating cache lines that the accelerator is going to read
and/or write, resulting in both costly performance overheads and a challenging
programming model where developers must manage complex coherency
operations. This has driven researchers to investigate alternative interfaces,
such as hardware-managed caches \cite{fusion, spandex, border_control,
crossing_guard, alsop2019optimizing}. While cache coherency for programmable accelerators like GPUs
have been extensively studied \cite{amd_compute_cores, nvidia_um,
gpu-tlb-abhishek, power2013, power2014, tabbakh2018g}, only in recent years have academia and
industry started investigating use of caches for fixed-function
accelerators and FPGAs. Full cache coherency represents the ideal
programming model, but the hardware is more expensive and generally
requires the accelerator to maintain a cache, which may not actually suit the
accelerated kernel.

In this case study, we explore a recent interface design that occupies a middle
ground between SW-managed and fully hardware-managed coherency. Here, the
interface provides one-way coherent access from the accelerator to the host
memory system. This interface takes the form of a special port, referred to as an accelerator coherency port (ACP), that issues coherent memory
requests directly to the CPU's last level cache (LLC). The LLC handles all
coherency traffic on the accelerator's  behalf. This enables the accelerator to
access coherent memory without adding more area and complexity.  To model such
an interface, we augment a standard MESI cache coherence protocol using the
Ruby modeling framework with a custom controller. This controller is connected
to an accelerator's memory interface and generates requests to the LLC on
behalf of the accelerator. Unlike a standard cache controller, it does not
implement a cache and leaves ownership of the relevant cache lines with the LLC
rather than the accelerator itself. Using Verilog simulation of an ARM Cortex
A53 CPU, we measure ACP hit latency to be 20 cycles, which we set as the LLC
latency.

Figure \ref{fig:soc-interface} shows the performance and energy of the ACP
interface, relative to the baseline DMA.  By attaching the DNN accelerator over
ACP, DNN performance improves by 17-55\% and energy consumption drops by
up to 56\%.  This is attributed to two effects. First, as a coherent interface, ACP
eliminates the software coherency management overhead associated with using DMA
for data transfers, which prior work \cite{shao_micro2016} has shown to be a significant fraction of
overall data transfer time. This accounts for the majority of the speedup on
data transfers.  Second, when using this coherent interface, many expensive
DRAM accesses are converted into cheaper LLC hits, which reduces overall energy
consumption by around 20\% on average, as shown in Figure
\ref{fig:soc-interface-energy}.  While the actual improvements vary based on
the size of the network and the tiling configurations, all these performance
and energy wins were achieved just by changing the interface, not the
accelerator. As the number of specialized blocks on SoCs increases over time,
optimizing interfacing choices will become increasingly important and challenging.  SMAUG enables
researchers to study these challenging system-level architecture choices using full-stack deep learning
workloads.

\subsection{Improving Accelerator Compute: Multi-Accelerator Systems}

\begin{figure}[t]
\centering
\includegraphics[width=8cm]{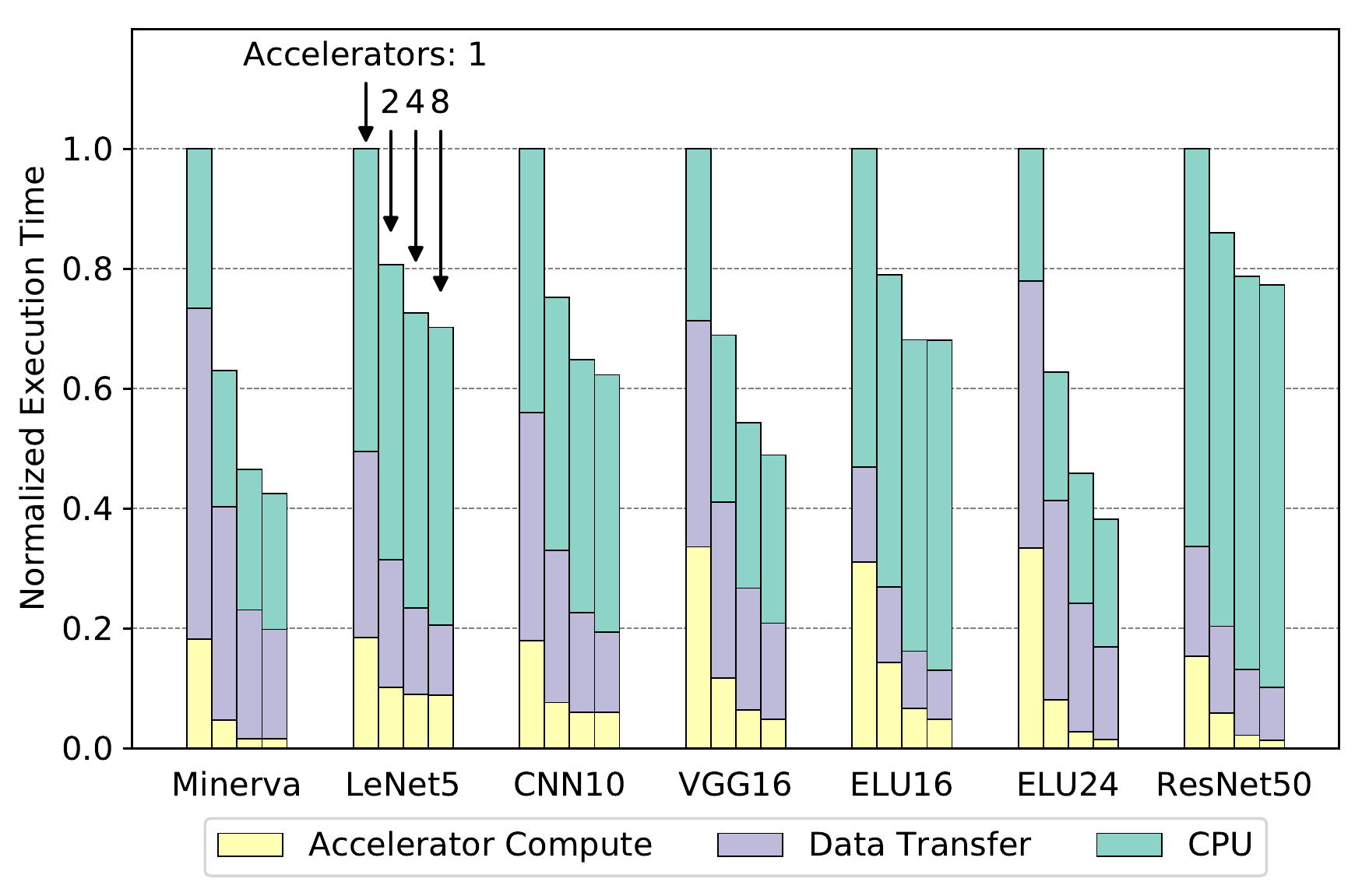}
\caption{Execution time of multiple-accelerator systems.}
\vspace{-0.4cm}
\label{fig:multiple-accel-cycles}
\end{figure}

DNN workloads have many different levels of parallelism, whether it's in the
parallel arithmetic operations within a tile, across tiles within a single
operation, or across independent operations entirely (like residual branches in
ResNet50). In this section, we explore scaling multi-accelerator systems to
better exploit tile-level parallelism.  Compared to a single monolithic block,
a multi-accelerator system (e.g. spatial arrays or multi-chip modules) with
independently programmable components can potentially scale to larger designs
more easily and be more flexible for different workloads. We choose to exploit
tile-level parallelism for two reasons: first, exploiting parallelism across
arithmetic operations lies at the intersection of finding better tiling shapes
for wider, more efficient accelerator datapaths, and second, it is a more
universal feature of DNNs compared to inter-operator parallelism (like residual
branches).

When multiple accelerators are available, SMAUG places them into an pool of
workers. Each accelerator is controlled directly by the runtime scheduler. When
tiling for a layer is finished, the scheduler pushes tiles of work to the
command queue of the assigned accelerator and tracks the progress of all tiles
in flight.  New tiles are pushed to the queue once their data dependencies are
resolved; for example, some tiling configurations need all partial products
along the channel dimension to be reduced before moving on to the next block of
rows or columns. However, dividing work across multiple accelerators is not
free, nor does it always improve performance. For example, if the dataflow is
input-stationary and requires each of the N accelerators to share a weight
tile, the weight data must now be broadcast to N destinations instead of just
1, which means extra data movement.

\begin{figure}[t]
\centering
\subfloat[Overall memory traffic for the entire workload.]{
	\label{fig:multiple-accel-traffic}
	\includegraphics[width=8cm]{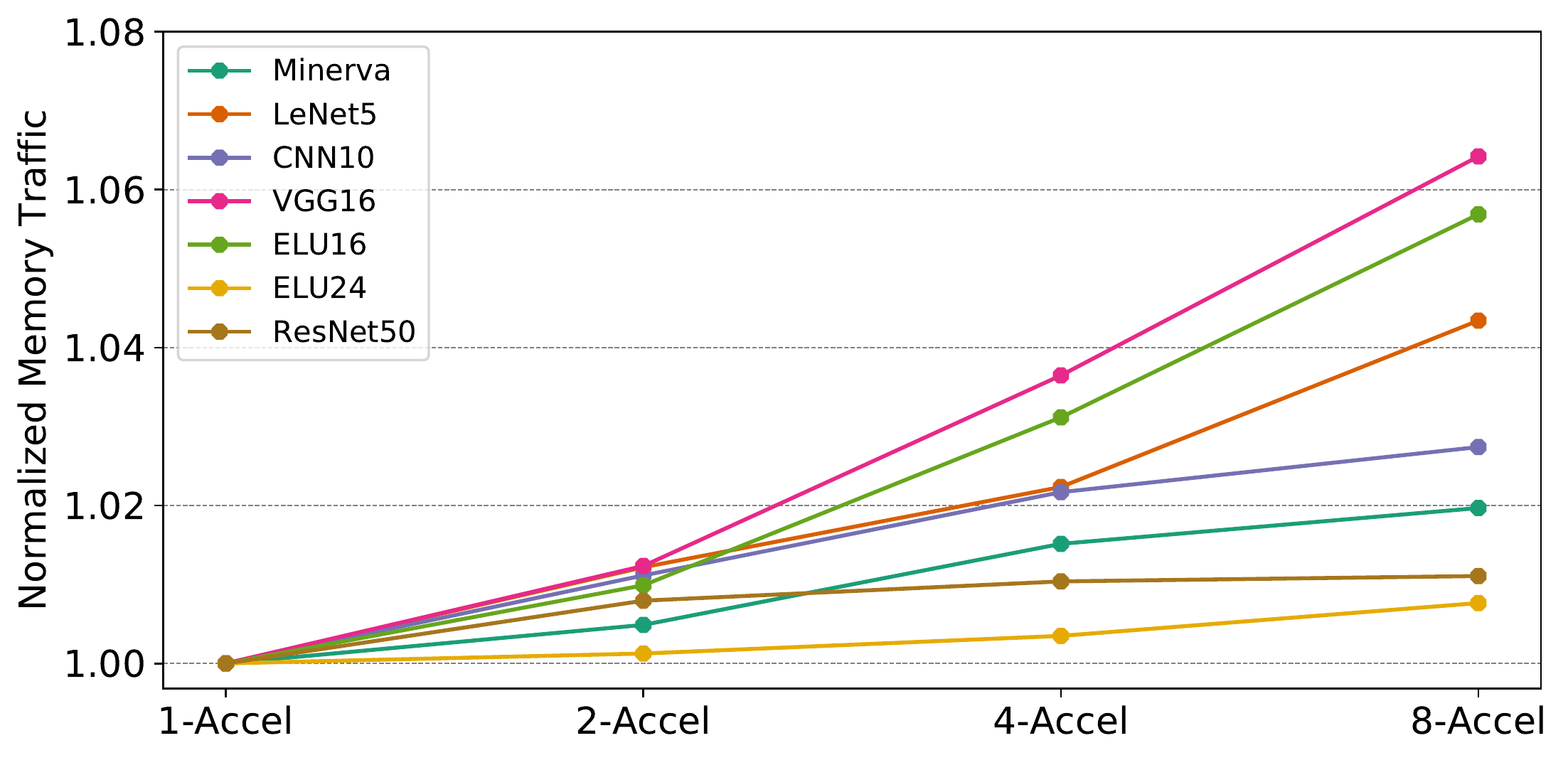}}
\hfill
\subfloat[Average end-to-end memory bandwidth utilization.]{
	\label{fig:multiple-accel-bandwidth}
	\includegraphics[width=7.9cm]{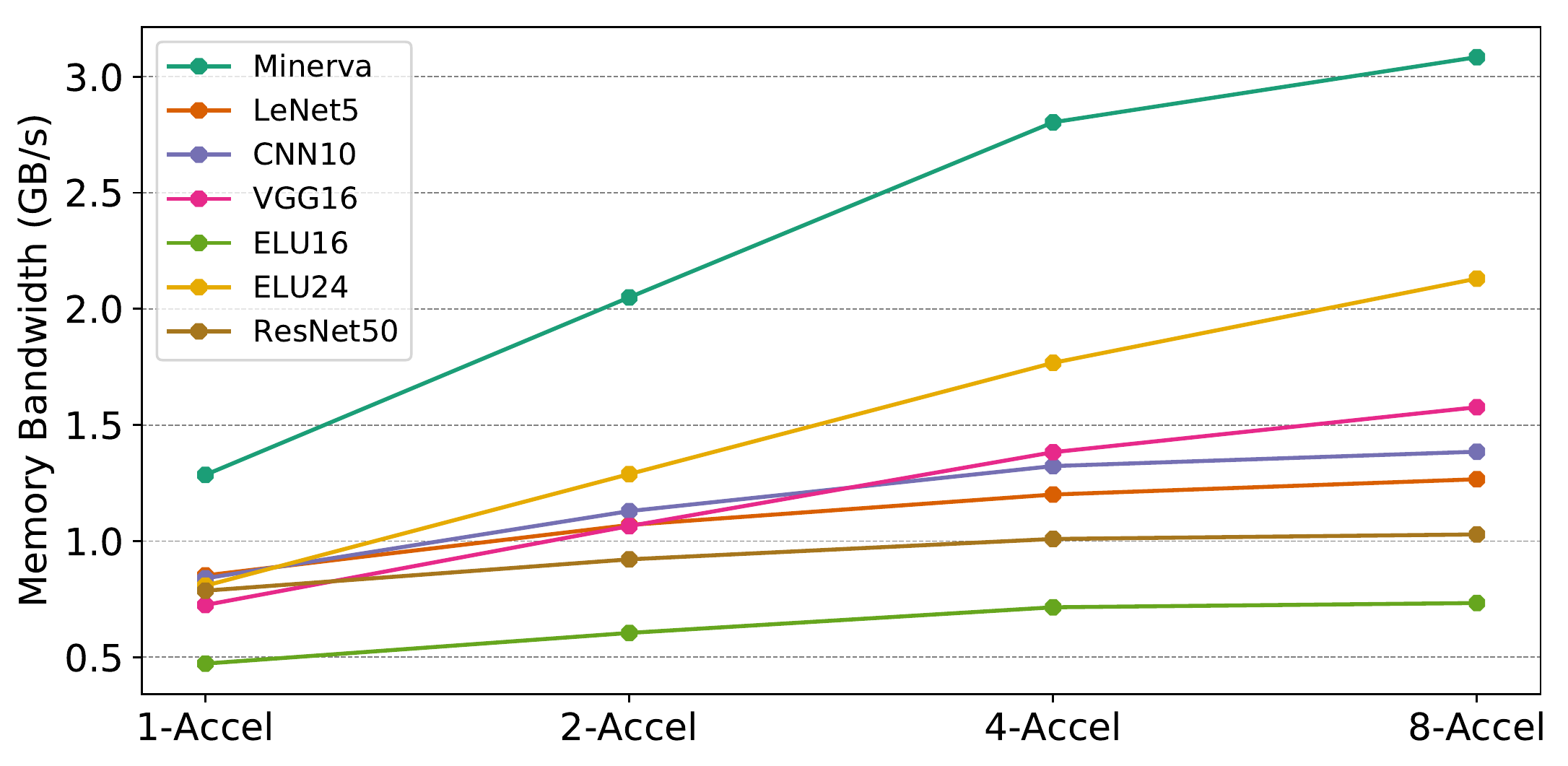}}
\caption{Memory traffic and bandwidth usage of multiple-accelerator systems, averaged
over the course of the entire workload.}
\vspace{-0.4cm}
\label{fig:multiple-accel-memory-stats}
\end{figure}

Figure \ref{fig:multiple-accel-cycles} shows how performance of
multi-accelerator systems scales with accelerator count.  As expected,
accelerator compute time speedup is consistent with the increase in available
processing units. It continues until we saturate the available tile-level
parallelism, which naturally occurs earlier for smaller networks than larger
ones.  Increasing accelerator count also means increasing total DRAM bytes
transferred because some data will need to be broadcast to all PEs, but 
as Figure \ref{fig:multiple-accel-traffic} shows, this effect is small
in the context of the entire workload, with at most a 6\% increase in overall traffic.
On the other hand, Figure \ref{fig:multiple-accel-bandwidth} shows that
multiple accelerators are also able to make better use of the available memory
bandwidth. Overall, data transfer time drops by around 60\% on
average. Together, with eight accelerators in the system, end-to-end latency
improves by between 20-60\% over a single-accelerator system.  This case study
demonstrates how SMAUG can clearly illuminate the overall performance
bottlenecks in DNN performance: by the time we reach eight accelerators,
compute time is negligible compared to data transfer time and software stack
time, and therefore those are the next components to optimize.




\begin{figure}[t]
\centering
\includegraphics[width=8cm]{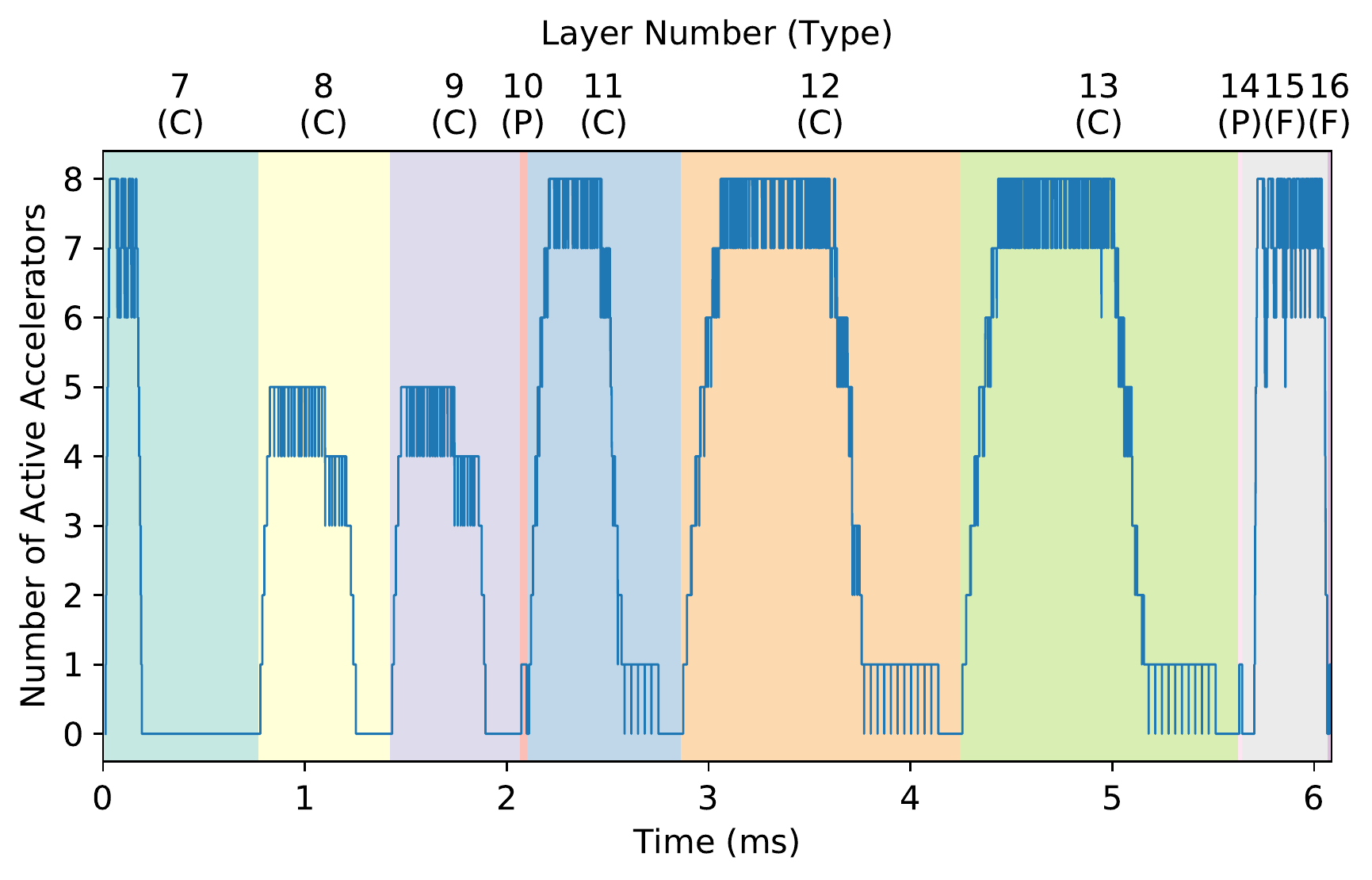}
\caption{Accelerator utilization of VGG16 with 8 accelerators. C, P and F stand for layer types of convolution, pooling and fully-connected, respectively.}
\label{fig:multiple-accel-load}
\vspace{-0.6cm}
\end{figure}

When debugging bottlenecks in DNN inference, it is useful to inspect
per-operation performance or performance between two particular operations.
With SMAUG, we can generate an execution timeline of important events for users
to visualize.  For example, Figure \ref{fig:multiple-accel-load} shows the
accelerator utilization for the last ten layers of VGG16, when the system has
eight accelerators in total. These
layers contain the largest six convolutional layers by number of parameters, 2
pooling layers (2x2), and 2 fully-connected layers (512 and 10 neurons each).
The timeline illustrates several opportunities for further optimization, which
we summarize below.

\textbf{Work balancing for higher accelerator utilization.}
The timeline shows that layers 8 and 9 are not fully utilizing all the
accelerators in the system because for this accelerator, the runtime scheduler
only supports in-place reduction of partial products along the channel dimension,
so all the tiles whose partial products must be reduced are put onto the same
accelerator's command queue. Then for this layer shape, there are only five
output tiles (i.e. independent streams of work), so only five accelerators are
used.  It is possible to evenly distribute work across all workers, which
would require the runtime scheduler to support inter-accelerator reduction.
Overall, the runtime scheduler in SMAUG does a good job in exploiting
tile-level parallelism in the DNN; we leave further optimization of the scheduler
for future work.

\textbf{Accelerating inter-layer tiling operations.}
The timeline shows that on Layer 7, the accelerator finishes computation very
quickly, followed by a long period of CPU activity. This is the CPU performing
``data finalization'': gathering all the output tiles from the accelerators and
rearranging them into a single tensor (effectively ``untiling'' the tensor),
because the next layer will likely need different input tile shapes.  Ways to
optimize this includes adjusting tiling shapes to maximize regions for
contiguous \texttt{memcpys} (see Figure \ref{fig:tiling-tradeoff}) and distributing the
work across multiple CPUs to increase task-level parallelism and memory
bandwidth utilization, which is the subject of the next section.

\subsection{Improving Software Stack: Multithreaded Data Management}

After all the effort spent optimizing core kernels like matrix-multiply and
convolution, the performance bottleneck shifts to the cost of preparing
data for use, and since this preparation is typically part of the
software framework, the overhead is exaggerated in comparison to the
accelerated kernels. This is not specific to SMAUG; on industry-grade
recommendation models, data preparation, other framework
native operations, and synchronization can take up anywhere from 10 to over
70\% of inference latency \cite{emmawang2019, park2018deep}.  In this case study, we look at
ways to reduce this overhead.

We break down the execution time of the software stack into three parts: data
preparation, data finalization, and other software activities. Data preparation
includes layout transformations, in which the dimensions of a tensor are either
rearranged (e.g. NCHW to NHWC) or flattened, and tensor tiling, which copies
non-contiguous logical regions of one tensor into contiguous smaller tensors
which can then be directly transferred to the accelerator for computation.  As
a result, when accelerators finish their work, their output tensors are also
tiled, which must now be ``untiled'' to obtain the final output tensor. We
refer to this untiling operation as data finalization.  Finally, other software
activities include tasks like control flow management, memory management,
various glue logic, and thread synchronization.

\begin{figure}[t]
\centering
\includegraphics[width=8cm]{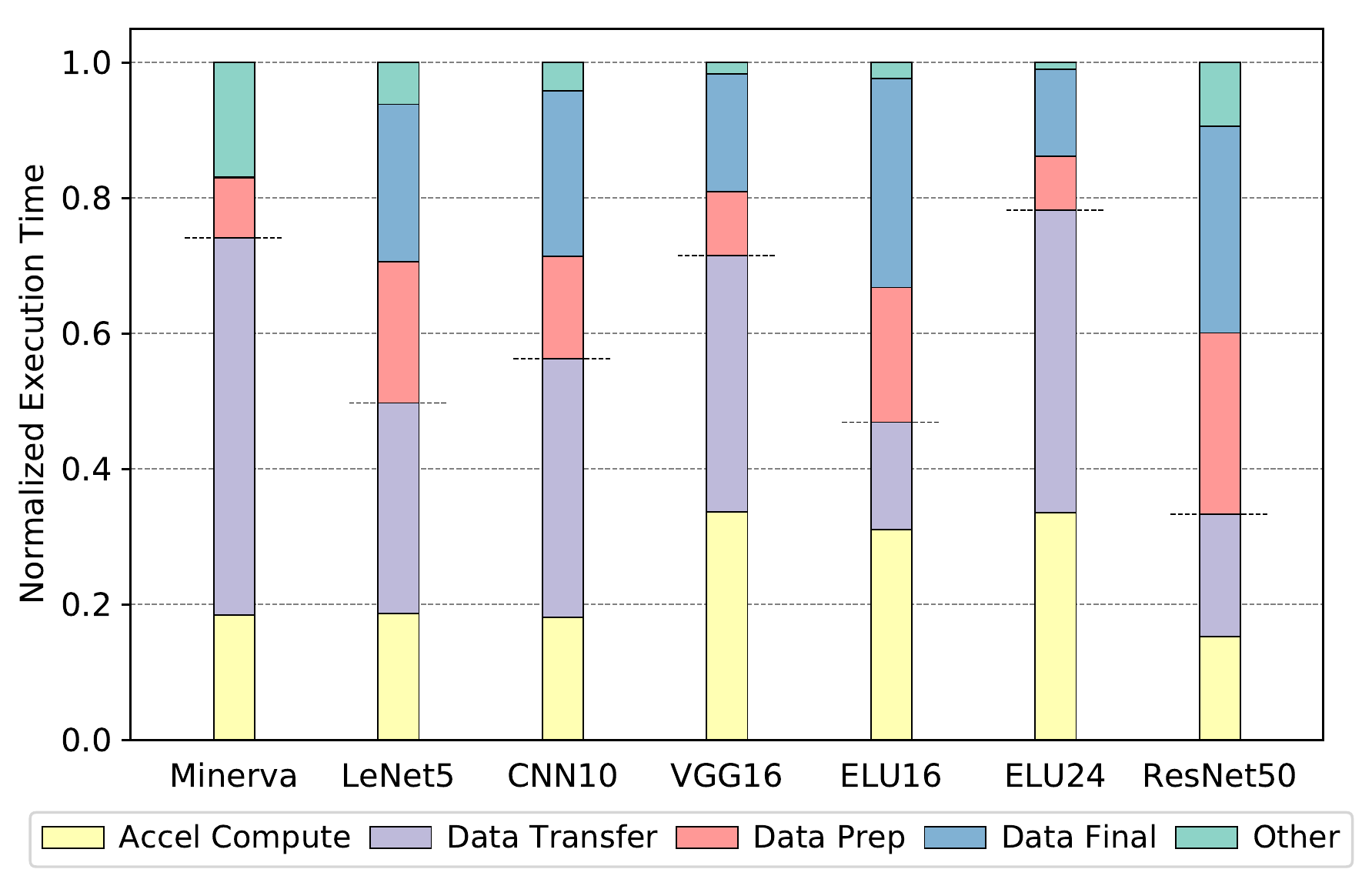}
\caption{Execution time breakdown of the baseline system. The dotted line separates hardware and software stack time.}
\label{fig:software-stack-baseline-cycles}
\end{figure}


\begin{figure}[t]
\centering
\includegraphics[width=8cm]{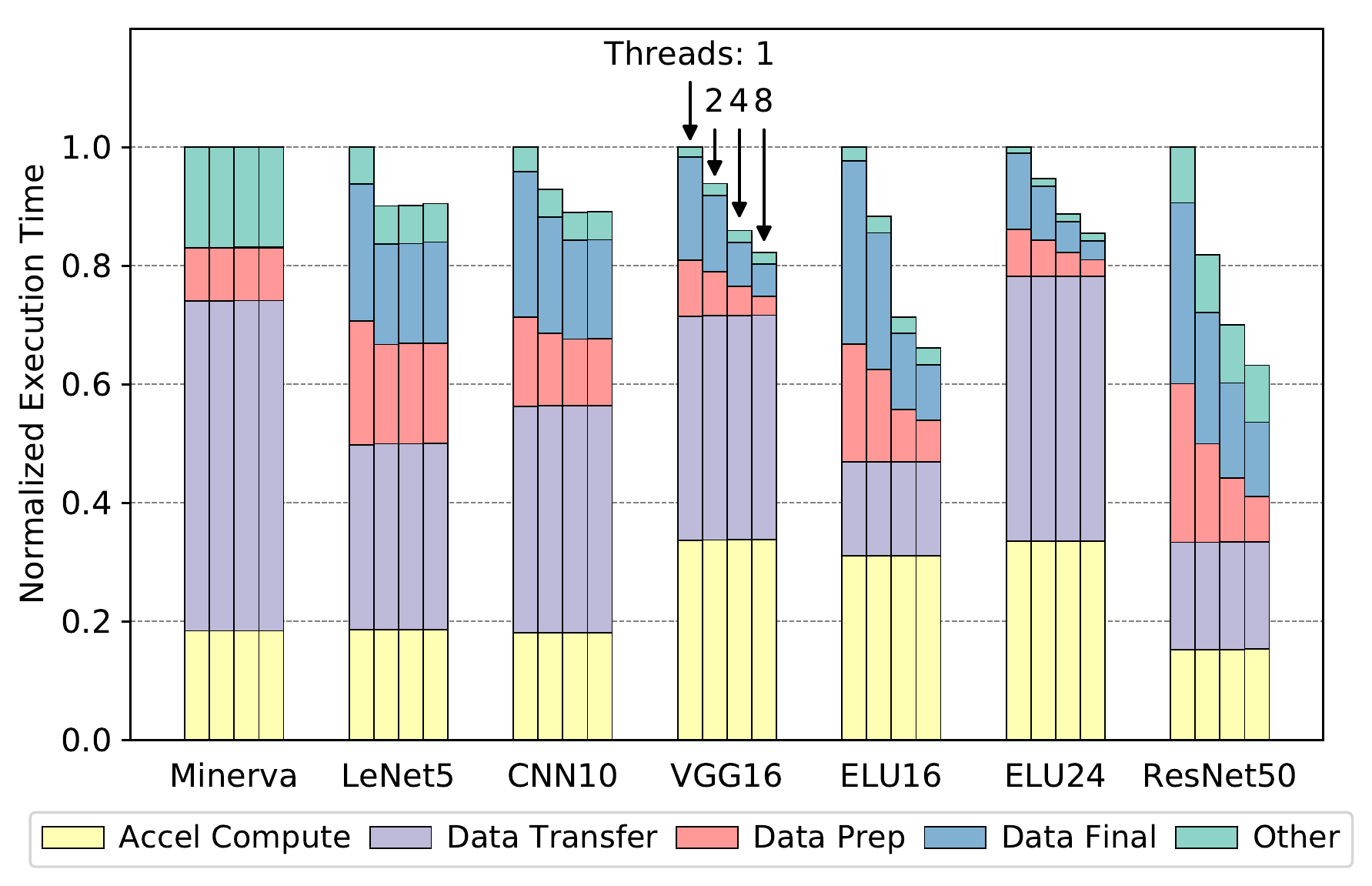}
\caption{Execution time of the system with a multithreaded software stack.}
\label{fig:software-stack-cycles}
\vspace{-0.4cm}
\end{figure}

\begin{figure}[t]
\centering
\includegraphics[width=8cm]{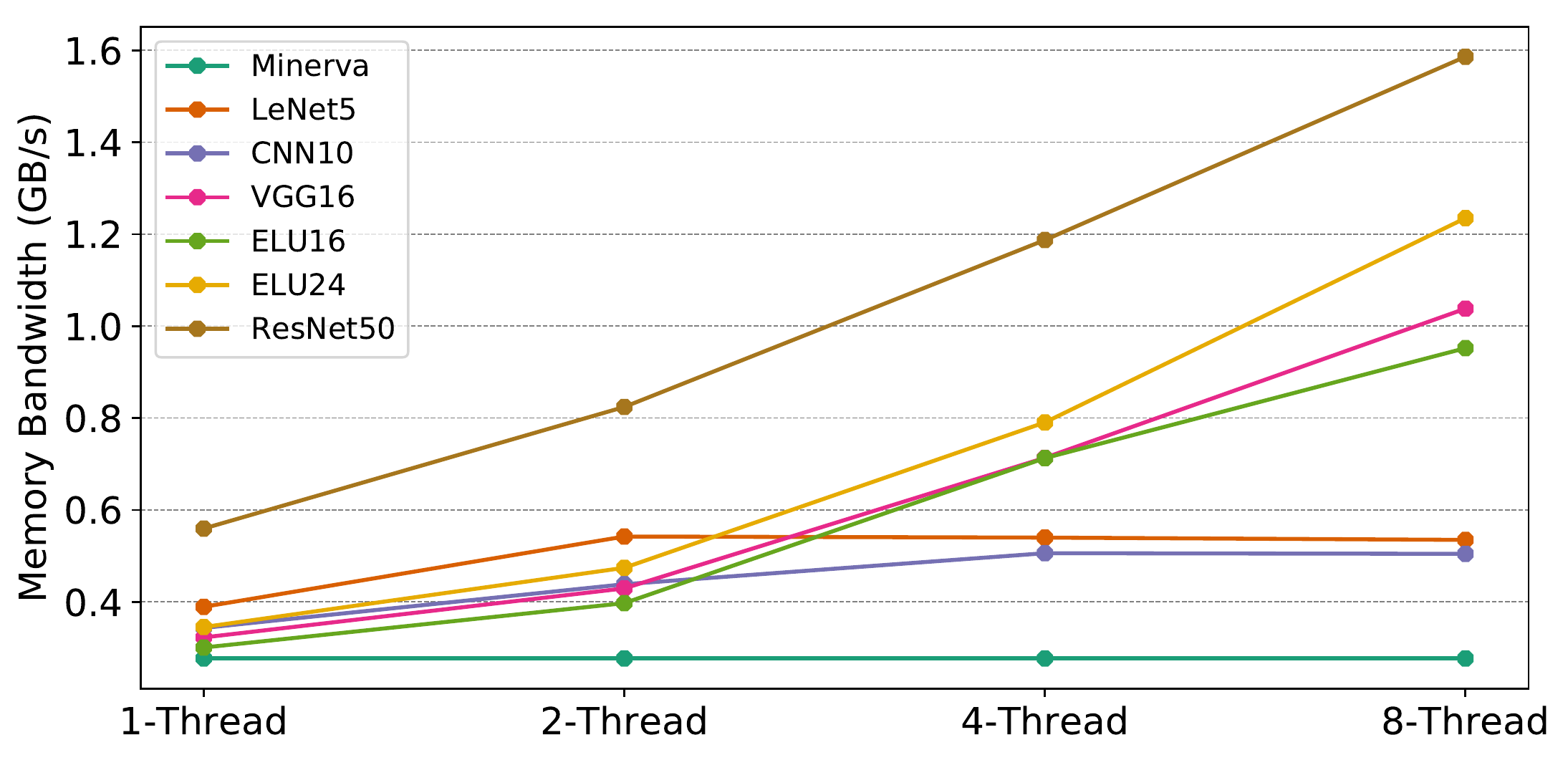}
\caption{Bandwidth utilization during the data preparation and gathering.}
\label{fig:software-stack-bandwidth}
\vspace{-0.2cm}
\end{figure}


Figure \ref{fig:software-stack-baseline-cycles} shows that on the baseline
system, data preparation and finalization account for 85\% of the software
stack time, so there is ample room for improvement.  As with the previous
section, we attack this problem through tile-level parallelism. We use SMAUG's
thread pool (see Section \ref{sec:multithreading}) to distribute data
preparation and finalization tasks across multiple threads.  Each thread is
responsible for copying data to/from a set of tiles. The baseline system has
already accounted for the cost of tiling transformations when determining the
best available tiling strategy.

With multithreaded tiling, we can achieve up to 3-$4\times$ speedup on data
preparation/finalization with eight threads, as shown in Figure
\ref{fig:software-stack-cycles}, resulting in an end-to-end latency reduction
of up to 37\%.  This speedup is primarily due to an increase in memory
bandwidth utilization when multiple threads are active.  Figure
\ref{fig:software-stack-bandwidth} shows the memory bandwidth usage during the
data preparation and gathering phases of the multithreaded software stack. On
large networks like ResNet50, which have a lot of tiles, multiple threads
increases bandwidth utilization by $2.7\times$ and leads to a $2.8\times$
speedup on data preparation and finalization tasks, while smaller networks like
Minerva don't have enough tile-level parallelism for multi-threading to
exploit.

\subsection{Overall Combined Speedup}

\begin{figure}[t]
\centering
\includegraphics[width=8cm]{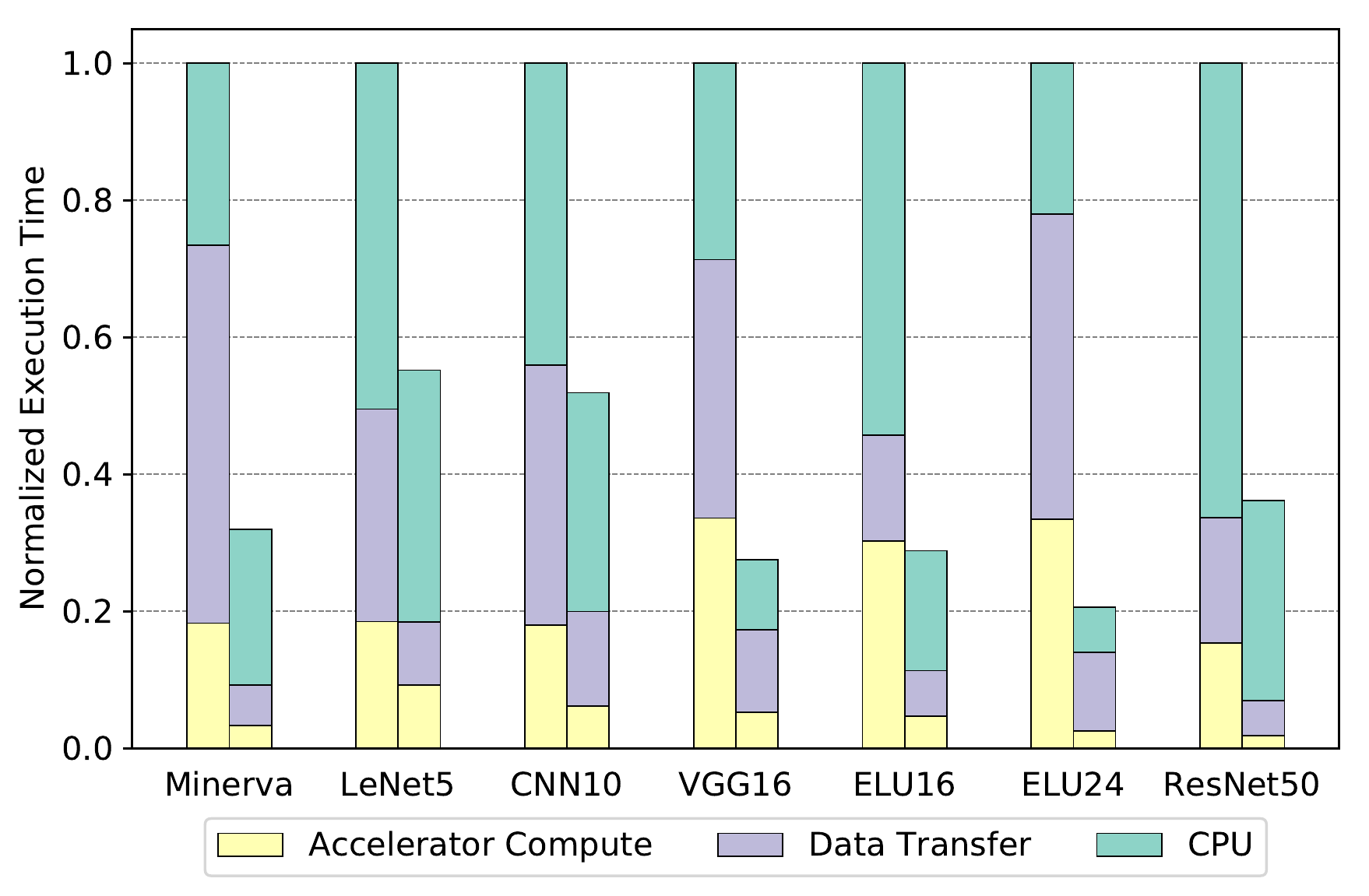}
\vspace{-0.3cm}
\caption{By combining all the optimizations to each component of performance from
  the case studies, we can reduce overall latency by 42\%-80\%.}
\label{fig:combined-cycle-breakdown}
\vspace{-0.4cm}
\end{figure}

Figure \ref{fig:combined-cycle-breakdown} summarizes the combined effect of the
three case studies on a single forward pass through all the networks.  The SoC
uses the ACP interface with eight accelerators and eight threads in the
software stack.  Overall latency drops by between 45\% on LeNet5 to as much as
80\% on ELU24 (1.8-$5\times$ speedup), all without changing any part of the
accelerator microarchitecture. This is a demonstration of the power of SMAUG
applied to system-level performance optimization of DNN workloads.

\section{Optimizing A Camera-Powered Deep Learning Pipeline}
\label{sec:case-study-camera}

\begin{figure*}[t]
\includegraphics[width=\textwidth]{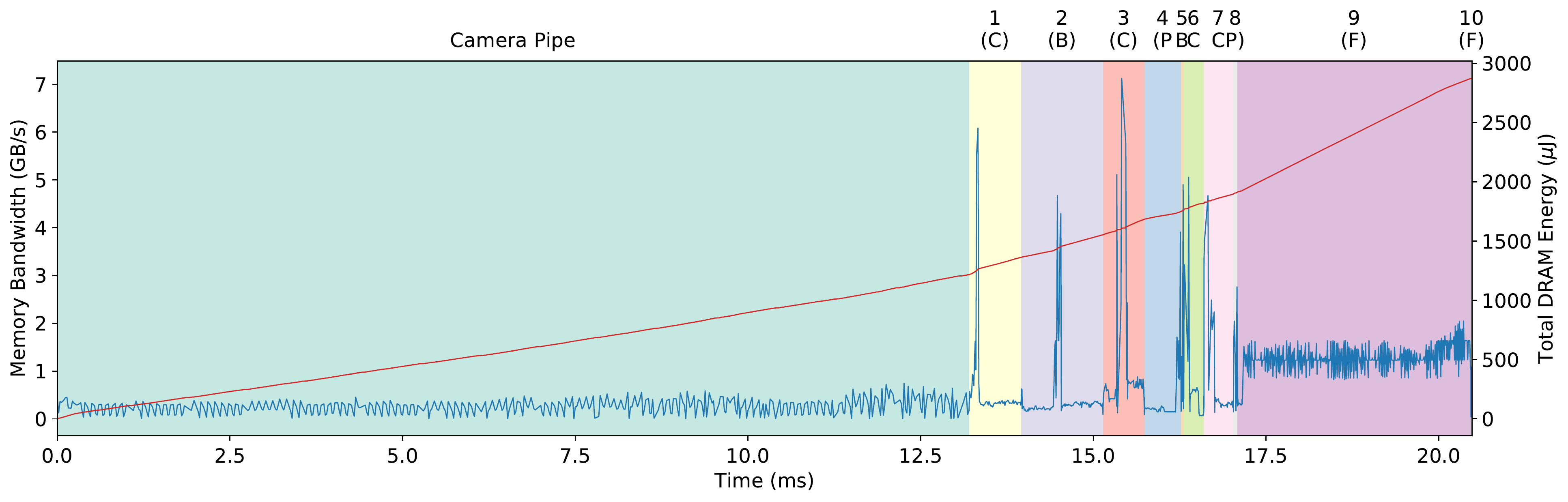}
\vspace{-0.5cm}
\caption{Execution trace of the camera vision pipeline running one frame. The
  vision part runs CNN10. C, B, P and F stand for layer types of convolution,
  batch normalization, pooling and fully-connected, respectively.}
\label{fig:camera-pipe-breakdown}
\vspace{-0.4cm}
\end{figure*}

In recent years, it has become increasingly common to attach deep learning
models at the end of other applications. One notable such application uses the
camera pipeline with a DNN to perform real-time tasks such as object classification, detection, segmentation and
labeling.  In this study, we
demonstrate how SMAUG can also model this kind of application and enable hardware-software 
co-design for better performance and energy efficiency.


The camera pipeline is a long series of spatial linear and non-linear filters
and transforms to convert the raw output of the image sensor into a realistic RGB representation. The sensor itself sits behind a Bayer color filter, so each individual
photodiode only captures light from one primary color (RGB). As a result, the
output of the sensor is an array of pixel values, each representing the
intensity of a single color. The process that estimates the original color of
each pixel from this raw image is called demosaicing. The subsequent image
processing then proceeds through many more processing steps, like white balance
correction, color space conversion, chroma subsampling, and more.  Finally, the
image is compressed in a lossy format (e.g. JPEG), which preserves low
frequency details that human eyes are sensitive to while removing imperceptible
high frequency information \cite{fatahalian2011}.

To construct such a camera vision pipeline, we integrate the complete camera
pipeline implementation shipped with Halide \cite{halide} into SMAUG and
simulate it as a single process running on the CPU. The camera pipeline transforms raw data
recorded by camera sensors into usable 720p images, including several stages:
hot pixel suppression, deinterleaving, demosaicing, white balancing and
sharpening. Modern image sensors use multi-megapixel resolutions, but that
resolution is often not necessary for DNNs, in this study, we feed 720p images
through the camera pipeline, then downsample it to the size required by the
DNN. For real-time applications, frame-time is a more representative metric of
responsiveness than throughput, so assuming the application targets 30 FPS
throughput, each frame must complete within 33 ms.  The baseline system
configuration we use is the same as the earlier case studies, except that to
show the accelerator variety in SMAUG, we use the
systolic array model (a cycle-level timing model written as a native gem5
object), configured as an 8x8 PE array instead of the NVDLA-inspired model.

\begin{figure}[t]
\centering
\includegraphics[width=0.47\textwidth]{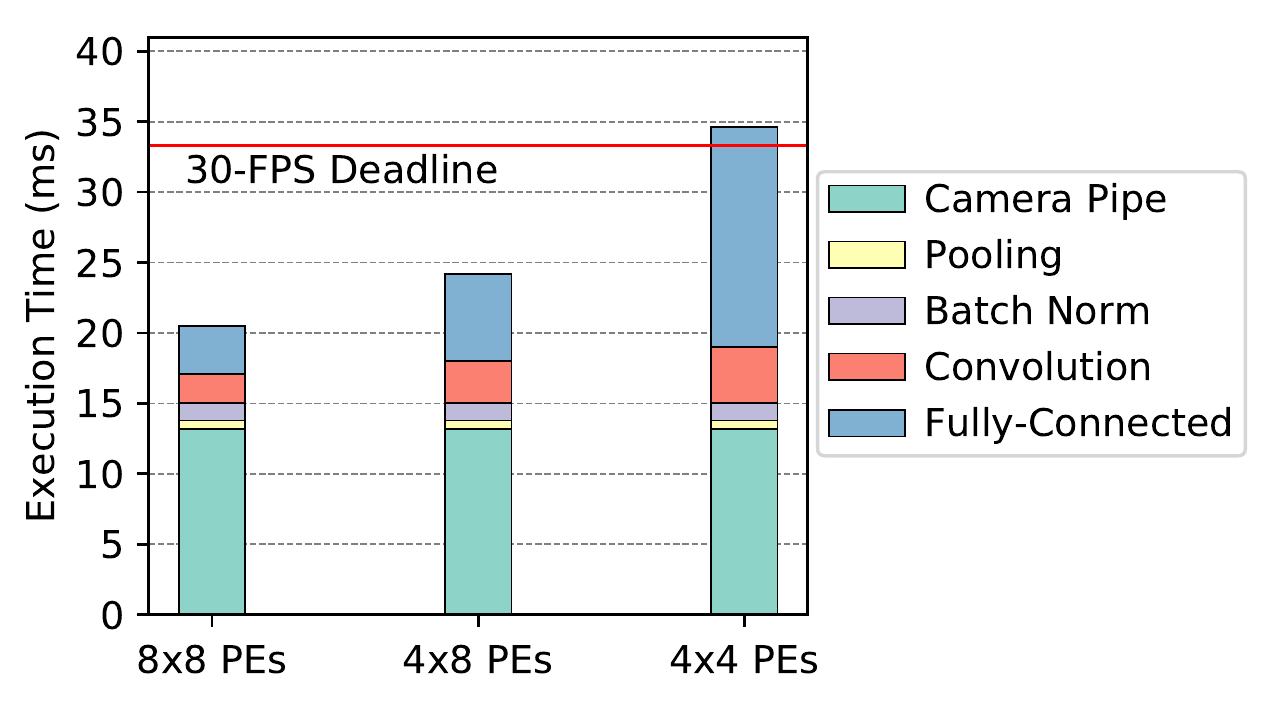}
\vspace{-0.4cm}
\caption{Execution time of the camera vision pipeline with different systolic array PE configurations.}
\vspace{-0.4cm}
\label{fig:camera-pipe-sweep}
\end{figure}

Figure \ref{fig:camera-pipe-breakdown} shows the execution timeline for the
camera vision pipeline, using the CNN10 network. With SMAUG, we can produce a
trace of memory bandwidth utilization and total memory energy consumed during
the application. In this case, the overall pipeline takes 20.5 ms to finish
(13.2ms of camera pipeline and 7.3ms of DNN), and memory energy consumption is
well balanced between the CPU (43\%) and accelerator (57\%).  The slack
time (12.8 ms) before the frame deadline means that in energy or chip area constrained
scenarios, we could afford to use an even smaller systolic array. 
As shown in Figure \ref{fig:camera-pipe-sweep}, reducing the PE array in half (4x8) increases the DNN latency to 11.0
ms, which still meets the frame-time limits. However, further decreasing the
PEs to a 4x4 array results in an overall latency of 34.6 ms, violating the
real-time constraint. Most of this extra latency comes from the final classifier
layer.

\section{Related Work}

\label{sec:related}

\begin{sloppypar}

\textbf{Simulation frameworks.}
SoC-accelerator simulators usually require the user to implement the accelerators in RTL or using HLS tools \cite{huangcentrifuge, parade, liang2017paas, piccolboni2017broadening, chidambaram2014gemdroid, platfrom_architect}. Centrifuge proposes a prototyping methodology that leverages HLS to generate accelerator SoCs and deploy them to FPGAs \cite{huangcentrifuge}. PARADE combines HLS with gem5 for full-system simulation \cite{parade}. GemDroid couples gem5 with the Android Emulator and integrates various hardware IP models to enable SoC-level simulation \cite{chidambaram2014gemdroid}. The heavy reliance on RTL implementation significantly increases the algorithm-to-solution time, even with HLS tools. In contrast, SMAUG builds on top of gem5-Aladdin, which uses a pre-RTL approach to accurately model the power, performance, and area of accelerator designs.

Table \ref{tab:infrastructure} lists a selection of deep learning research frameworks. Some are
end-to-end systems, like TensorFlow \cite{abadi2016tensorflow} or TVM \cite{TVM}, but they either do not support simulation or require detailed pipeline models or RTL.
Other tools focus on exploring dataflows and
efficiently  map DNN kernels to FPGAS or ASICs \cite{zhang2015optimizing,sharma2016dnnweaver, zhang2018dnnbuilder, venkatesanmagnet, timeloop, wu2019accelergy}. These often implement a component library or optimized template designs for hardware optimization, but
with a heavy focus on optimizing the accelerator, they can't evaluate networks end-to-end, 
leaving a lot of design opportunities unexplored.
While all of these tools have their place in the deep learning research infrastructure landscape, 
they do not provide SMAUG's capability of enabling end-to-end early-stage design space exploration of the
SoC as well as the accelerator.


Due to the regularity of DNNs, there are simulation tools that apply analytical models for DNN performance analysis \cite{scale-sim, platfrom_architect}. SCALE-Sim \cite{scale-sim} models the systolic array accelerator with a variety of dataflows, each based on a different analytical model. While analytical models are fast for performance estimation, they cannot capture dynamic runtime behavior. SMAUG, however, captures both accelerator and system level activities while maintaining fast simulation speed with the accurate sampling support.

\textbf{DNN Accelerator Designs.}
There has been an incredible amount of interest in DNN hardware acceleration.
Broadly speaking, the architecture community has focused on designing efficient
dataflows to maximize local reuse of data and functional unit
utilization~\cite{googletpu, chi2016prime, liu2016cambricon,ding2017circnn,
eyeriss, alwani2016fused, lu2017flexflow, kwon2018maeri}, exploit model sparsity and data
quantization \cite{qin2019, zhu2019micro, wang2019hpca, judd2016stripes, scnn, eie, minerva}, 
map DNN accelerators to FPGAs~\cite{FPDNN, DLAU, MemManagementFPGA}, explore
alternative computation and memory technologies \cite{shafiee2016isaac, wang2018dac}, or use multi-chip-module package 
integration to achieve high-performance DNN inference~\cite{simba}.  Although these
works are highly relevant, these papers do not address end-to-end performance
evaluation or between-the-layer operations, like data layout transformations.

\textbf{SoC-Accelerator Interfacing.}
Over the years, there have been a few publications investigating
SoC-accelerator interfacing and interactions in a variety of contexts, 
such as CoRAMs \cite{chungfpga2011}, $\mu$Layer \cite{ulayer2019}, and
Google mobile system workloads \cite{boroumand2018}.
A few recent works have considered
interfacing between the SoC and accelerators \cite{zhu2018vision, eva2, asv}.
A handful of other works have used the ARM accelerator coherency port for
tighter coupling between CPU and accelerators, albeit not in the context of
DNNs \cite{sadri_fpgaworld2013, moreau2015snnap}.


\end{sloppypar}

\section{Acknowledgement}
This work was supported in part by the U.S. Government, under the
DARPA DSSoC program. This work was also supported in part by the
Semiconductor Research Corporation, NSF grant $\#$ CNS-1718160, and Intel.
\section{Conclusion}
\label{sec:conclusion}
This paper demonstrates the critical importance of evaluating full-stack performance of a hardware accelerated computing task
like neural network inference. Recent years have brought great advances in accelerator design and efficient DNN dataflows,
but several important components of overall performance, like data transformation and movement cost and software
framework overheads, have received far less attention, partly because of a lack of suitable research infrastructure.
We developed SMAUG, a DNN framework that can be simulated in a cycle-level SoC simulator, and demonstrate how 
it can be used to optimize end-to-end performance on a wide range of DNNs to achieve between 1.8-5$\times$ speedup by 
optimizing SoC-accelerator interfaces, exploiting multi-accelerator systems, and optimizing the software stack.
Since SMAUG provides architects with a straightforward approach to simulate complex full-stack workloads, we hope it will spur renewed interest in broader optimization of end-to-end performance in DNN hardware studies.

\bibliographystyle{IEEEtran}
\bibliography{ref}

\begin{thebibliography}{10}
\providecommand{\url}[1]{#1}
\csname url@samestyle\endcsname
\providecommand{\newblock}{\relax}
\providecommand{\bibinfo}[2]{#2}
\providecommand{\BIBentrySTDinterwordspacing}{\spaceskip=0pt\relax}
\providecommand{\BIBentryALTinterwordstretchfactor}{4}
\providecommand{\BIBentryALTinterwordspacing}{\spaceskip=\fontdimen2\font plus
\BIBentryALTinterwordstretchfactor\fontdimen3\font minus
  \fontdimen4\font\relax}
\providecommand{\BIBforeignlanguage}[2]{{%
\expandafter\ifx\csname l@#1\endcsname\relax
\typeout{** WARNING: IEEEtran.bst: No hyphenation pattern has been}%
\typeout{** loaded for the language `#1'. Using the pattern for}%
\typeout{** the default language instead.}%
\else
\language=\csname l@#1\endcsname
\fi
#2}}
\providecommand{\BIBdecl}{\relax}
\BIBdecl

\bibitem{googletpu}
N.~P. Jouppi, C.~Young, N.~Patil, D.~Patterson, G.~Agrawal, R.~Bajwa, S.~Bates,
  S.~Bhatia, N.~Boden, A.~Borchers \emph{et~al.}, ``In-datacenter performance
  analysis of a tensor processing unit,'' in \emph{ISCA}, 2017.

\bibitem{eyeriss}
Y.-H. Chen, T.~Krishna, J.~Emer, and V.~Sze, ``{Eyeriss: An Energy-Efficient
  Reconfigurable Accelerator for Deep Convolutional Neural Networks},'' in
  \emph{{ISSCC}}, {2016}.

\bibitem{judd2016stripes}
P.~Judd, J.~Albericio, T.~Hetherington, T.~M. Aamodt, and A.~Moshovos,
  ``Stripes: Bit-serial deep neural network computing,'' in \emph{MICRO}, 2016.

\bibitem{chi2016prime}
P.~Chi, S.~Li, C.~Xu, T.~Zhang, J.~Zhao, Y.~Liu, Y.~Wang, and Y.~Xie, ``Prime:
  a novel processing-in-memory architecture for neural network computation in
  reram-based main memory,'' in \emph{ISCA}, 2016.

\bibitem{liu2016cambricon}
S.~Liu, Z.~Du, J.~Tao, D.~Han, T.~Luo, Y.~Xie, Y.~Chen, and T.~Chen,
  ``Cambricon: An instruction set architecture for neural networks,'' in
  \emph{ISCA}, 2016.

\bibitem{alwani2016fused}
M.~Alwani, H.~Chen, M.~Ferdman, and P.~Milder, ``Fused-layer cnn
  accelerators,'' in \emph{MICRO}, 2016.

\bibitem{ding2017circnn}
C.~Ding, S.~Liao, Y.~Wang, Z.~Li, N.~Liu, Y.~Zhuo, C.~Wang, X.~Qian, Y.~Bai,
  G.~Yuan \emph{et~al.}, ``Circnn: accelerating and compressing deep neural
  networks using block-circulant weight matrices,'' in \emph{MICRO}, 2017.

\bibitem{emmawang2019}
Y.~E. Wang, C.-J. Wu, X.~Wang, K.~Hazelwood, and D.~Brooks, ``Exploiting
  parallelism opportunities with deep learning frameworks,'' 2019.

\bibitem{park2018deep}
J.~Park, M.~Naumov, P.~Basu, S.~Deng, A.~Kalaiah, D.~Khudia, J.~Law, P.~Malani,
  A.~Malevich, S.~Nadathur, J.~Pino, M.~Schatz, A.~Sidorov, V.~Sivakumar,
  A.~Tulloch, X.~Wang, Y.~Wu, H.~Yuen, U.~Diril, D.~Dzhulgakov, K.~Hazelwood,
  B.~Jia, Y.~Jia, L.~Qiao, V.~Rao, N.~Rotem, S.~Yoo, and M.~Smelyanskiy, ``Deep
  learning inference in facebook data centers: Characterization, performance
  optimizations and hardware implications,'' 2018.

\bibitem{abadi2016tensorflow}
M.~Abadi, P.~Barham, J.~Chen, Z.~Chen, A.~Davis, J.~Dean, M.~Devin,
  S.~Ghemawat, G.~Irving, M.~Isard \emph{et~al.}, ``Tensorflow: A system for
  large-scale machine learning,'' in \emph{OSDI}, 2016.

\bibitem{pytorch}
N.~Ketkar, ``Introduction to pytorch,'' in \emph{Deep learning with
  python}.\hskip 1em plus 0.5em minus 0.4em\relax Springer, 2017.

\bibitem{caffe}
Y.~Jia, E.~Shelhamer, J.~Donahue, S.~Karayev, J.~Long, R.~Girshick,
  S.~Guadarrama, and T.~Darrell, ``Caffe: Convolutional architecture for fast
  feature embedding,'' in \emph{Proceedings of the 22nd ACM international
  conference on Multimedia}, 2014.

\bibitem{mxnet}
T.~Chen, M.~Li, Y.~Li, M.~Lin, N.~Wang, M.~Wang, T.~Xiao, B.~Xu, C.~Zhang, and
  Z.~Zhang, ``Mxnet: A flexible and efficient machine learning library for
  heterogeneous distributed systems,'' \emph{arXiv preprint arXiv:1512.01274},
  2015.

\bibitem{sharma2016dnnweaver}
H.~Sharma, J.~Park, E.~Amaro, B.~Thwaites, P.~Kotha, A.~Gupta, J.~K. Kim,
  A.~Mishra, and H.~Esmaeilzadeh, ``Dnnweaver: From high-level deep network
  models to fpga acceleration,'' in \emph{the Workshop on Cognitive
  Architectures}, 2016.

\bibitem{zhang2018dnnbuilder}
X.~Zhang, J.~Wang, C.~Zhu, Y.~Lin, J.~Xiong, W.-m. Hwu, and D.~Chen,
  ``Dnnbuilder: an automated tool for building high-performance dnn hardware
  accelerators for fpgas,'' in \emph{Proceedings of the International
  Conference on Computer-Aided Design}.\hskip 1em plus 0.5em minus 0.4em\relax
  ACM, 2018.

\bibitem{venkatesanmagnet}
R.~Venkatesan, Y.~S. Shao, M.~Wang, J.~Clemons, S.~Dai, M.~Fojtik, B.~Keller,
  A.~Klinefelter, N.~Pinckney, P.~Raina \emph{et~al.}, ``Magnet: A modular
  accelerator generator for neural networks,'' in \emph{{ICCAD}}, 2019.

\bibitem{TVM}
T.~Chen, T.~Moreau, Z.~Jiang, L.~Zheng, E.~Q. Yan, H.~Shen, M.~Cowan, L.~Wang,
  Y.~Hu, L.~Ceze, C.~Guestrin, and A.~Krishnamurthy, ``{TVM:} an automated
  end-to-end optimizing compiler for deep learning,'' in \emph{USENIX OSDI},
  2018.

\bibitem{scale-sim}
\BIBentryALTinterwordspacing
A.~Samajdar, Y.~Zhu, P.~N. Whatmough, M.~Mattina, and T.~Krishna,
  ``{SCALE}-sim: Systolic {CNN} accelerator,'' \emph{CoRR}, 2018. [Online].
  Available: \url{http://arxiv.org/abs/1811.02883}
\BIBentrySTDinterwordspacing

\bibitem{sun2019hsim}
M.~Sun, P.~Zhao, Y.~Wang, N.~Chang, and X.~Lin, ``Hsim-dnn: Hardware simulator
  for computation-, storage-and power-efficient deep neural networks,'' in
  \emph{Proceedings of the 2019 on Great Lakes Symposium on VLSI}.\hskip 1em
  plus 0.5em minus 0.4em\relax ACM, 2019.

\bibitem{shao_micro2016}
Y.~S. Shao, S.~L. Xi, V.~Srinivasan, G.-Y. Wei, and D.~Brooks, ``Co-designing
  accelerators and soc interfaces using gem5-aladdin,'' in \emph{MICRO}, 2016.

\bibitem{minerva}
B.~Reagen, P.~Whatmough, R.~Adolf, S.~Rama, H.~Lee, S.~K. Lee, J.~M.
  Hernandez-Lobato, G.-Y. Wei, and D.~Brooks, ``{Minerva: Enabling Low-Power,
  Highly-Accurate Deep Neural Network Accelerators},'' in \emph{{ISCA}}, 2016.

\bibitem{lecun_ieee98}
{Y LeCun and L. Bottou and Y. Bengio and P. Haffner}, ``{Gradient-based
  learning applied to document recognition},'' in \emph{{Proceedings of the
  IEEE}}, {1998}.

\bibitem{cifar10_vgg}
\BIBentryALTinterwordspacing
{Sergey Zagoruyko}, ``{Torch Blog: 92.45 on CIFAR10 in Torch},'' {2015}.
  [Online]. Available: \url{{https://torch.ch/blog/07/30/cifar.html}}
\BIBentrySTDinterwordspacing

\bibitem{clevert2013elu}
S.~H. Djork-Arne~Clevert, Thomas~Underthiner, ``Fast and accurate deep network
  learning by exponential linear units (elus),'' in \emph{ICLR}, 2016.

\bibitem{resnet}
K.~He, X.~Zhang, S.~Ren, and J.~Sun, ``Deep residual learning for image
  recognition,'' in \emph{Proceedings of the IEEE conference on computer vision
  and pattern recognition}, 2016.

\bibitem{reddi2019mlperf}
V.~J. Reddi, C.~Cheng, D.~Kanter, P.~Mattson, G.~Schmuelling, C.-J. Wu,
  B.~Anderson, M.~Breughe, M.~Charlebois, W.~Chou, R.~Chukka, C.~Coleman,
  S.~Davis, P.~Deng, G.~Diamos, J.~Duke, D.~Fick, J.~S. Gardner, I.~Hubara,
  S.~Idgunji, T.~B. Jablin, J.~Jiao, T.~S. John, P.~Kanwar, D.~Lee, J.~Liao,
  A.~Lokhmotov, F.~Massa, P.~Meng, P.~Micikevicius, C.~Osborne, G.~Pekhimenko,
  A.~T.~R. Rajan, D.~Sequeira, A.~Sirasao, F.~Sun, H.~Tang, M.~Thomson, F.~Wei,
  E.~Wu, L.~Xu, K.~Yamada, B.~Yu, G.~Yuan, A.~Zhong, P.~Zhang, and Y.~Zhou,
  ``Mlperf inference benchmark,'' 2019.

\bibitem{jouppi2015cacti}
{Norman P. Jouppi and Andrew B. Kahng and Naveen Muralimanohar and Vaishav
  Srinivas}, ``{CACTI-IO: CACTI with Off-Chip Power-Area-Timing Models},''
  \emph{{IEEE Transactions on VLSI Systems}}, vol.~{23}, {2015}.

\bibitem{drampower}
{Karthik Chandrasekar, Christian Weis, Yonghui Li, Sven Goossens, Matthias
  Jung, Omar Naji, Benny Akesson, Norbert Wehn, and Kees Goossens},
  ``{DRAMPower: Open-Source DRAM Power and Energy Estimation Tool},''
  {http://drampower.info}.

\bibitem{micron_lpddr4}
\emph{{Mobile LPDDR4 SDRAM}}, {Micron Technology}, {2014}.

\bibitem{fusion}
S.~Kumar, A.~Shriraman, and N.~Vedula, ``{Fusion: Design Tradeoffs in Coherent
  Cache Hierarchies for Accelerators},'' in \emph{ISCA}, 2015.

\bibitem{spandex}
J.~Alsop, M.~D. Sinclair, and S.~V. Adve, ``Spandex: A flexible interface for
  efficient heterogeneous coherence,'' \emph{ISCA}, 2018.

\bibitem{border_control}
L.~E. Olson, J.~Power, M.~D. Hill, and D.~A. Wood, ``{Border Control:
  Sandboxing Accelerators},'' in \emph{MICRO}, 2015.

\bibitem{crossing_guard}
L.~E. Olson, M.~D. Hill, and D.~A. Wood, ``Crossing guard: Mediating
  host-accelerator coherence interactions,'' \emph{ASPLOS}, 2017.

\bibitem{alsop2019optimizing}
J.~Alsop, M.~D. Sinclair, S.~Bharadwaj, A.~Dutu, A.~Gutierrez, O.~Kayiran,
  M.~LeBeane, S.~Puthoor, X.~Zhang, T.~T. Yeh \emph{et~al.}, ``Optimizing gpu
  cache policies for mi workloads,'' \emph{arXiv preprint arXiv:1910.00134},
  2019.

\bibitem{amd_compute_cores}
\BIBentryALTinterwordspacing
AMD, ``{Compute Cores},'' Tech. Rep., 2014. [Online]. Available:
  \url{www.amd.com/computecores}
\BIBentrySTDinterwordspacing

\bibitem{nvidia_um}
\BIBentryALTinterwordspacing
M.~Harris, ``{Unified Memory in CUDA 6},'' 2013. [Online]. Available:
  \url{https://devblogs.nvidia.com/parallelforall/unified-memory-in-cuda-6/}
\BIBentrySTDinterwordspacing

\bibitem{gpu-tlb-abhishek}
B.~Pichai, L.~Hsu, and A.~Bhattacharjee, ``Architectural support for address
  translation on gpus designing memory management units for cpu/gpus with
  unified address spaces,'' in \emph{ASPLOS}, 2014.

\bibitem{power2013}
J.~Power, A.~Basu, J.~Gu, S.~Puthoor, B.~M. Beckmann, M.~D. Hill, S.~K.
  Reinhardt, and D.~A. Wood, ``{Heterogeneous System Coherence for Integrated
  CPU-GPU Systems},'' in \emph{{MICRO}}, 2013.

\bibitem{power2014}
J.~Power, M.~D. Hill, and D.~A. Wood, ``Supporting x86-64 address translation
  for 100s of gpu lanes,'' in \emph{HPCA}, 2014.

\bibitem{tabbakh2018g}
A.~Tabbakh, X.~Qian, and M.~Annavaram, ``G-tsc: Timestamp based coherence for
  gpus,'' in \emph{2018 IEEE International Symposium on High Performance
  Computer Architecture (HPCA)}.\hskip 1em plus 0.5em minus 0.4em\relax IEEE,
  2018.

\bibitem{fatahalian2011}
\BIBentryALTinterwordspacing
K.~Fatahalian, ``A camera image processing pipeline,'' 2011. [Online].
  Available:
  \url{http://www.cs.cmu.edu/afs/cs/academic/class/15869-f11/www/lectures/16_camerapipeline1.pdf}
\BIBentrySTDinterwordspacing

\bibitem{halide}
J.~Ragan-Kelley, C.~Barnes, A.~Adams, S.~Paris, F.~Durand, and S.~Amarasinghe,
  ``Halide: a language and compiler for optimizing parallelism, locality, and
  recomputation in image processing pipelines,'' \emph{ACM SIGPLAN Notices},
  2013.

\bibitem{huangcentrifuge}
Q.~Huang, C.~Yarp, S.~Karandikar, N.~Pemberton, B.~Brock, L.~Ma, G.~Dai,
  R.~Quitt, K.~Asanovic, and J.~Wawrzynek, ``Centrifuge: Evaluating full-system
  hls-generated heterogeneous-accelerator socs using fpga-acceleration,'' in
  \emph{{ICCAD}}, 2019.

\bibitem{parade}
J.~Cong, Z.~Fang, M.~Gill, and G.~Reinman, ``{PARADE: A Cycle-Accurate
  Full-System Simulation Platform for Accelerator-Rich Architectural Design and
  Exploration},'' in \emph{{ICCAD}}, 2015.

\bibitem{liang2017paas}
T.~Liang, L.~Feng, S.~Sinha, and W.~Zhang, ``Paas: A system level simulator for
  heterogeneous computing architectures,'' in \emph{2017 27th International
  Conference on Field Programmable Logic and Applications (FPL)}.\hskip 1em
  plus 0.5em minus 0.4em\relax IEEE, 2017.

\bibitem{piccolboni2017broadening}
L.~Piccolboni, P.~Mantovani, G.~Di~Guglielmo, and L.~P. Carloni, ``Broadening
  the exploration of the accelerator design space in embedded scalable
  platforms,'' in \emph{2017 IEEE High Performance Extreme Computing Conference
  (HPEC)}.\hskip 1em plus 0.5em minus 0.4em\relax IEEE, 2017.

\bibitem{chidambaram2014gemdroid}
N.~Chidambaram~Nachiappan, P.~Yedlapalli, N.~Soundararajan, M.~T. Kandemir,
  A.~Sivasubramaniam, and C.~R. Das, ``Gemdroid: a framework to evaluate mobile
  platforms,'' \emph{ACM SIGMETRICS Performance Evaluation Review}, vol.~42,
  2014.

\bibitem{platfrom_architect}
{Synopsys Inc.}, ``{Synopsys Platform Architect},''
  \url{https://www.synopsys.com/verification/virtual-prototyping/platform-architect.html/}.

\bibitem{zhang2015optimizing}
C.~Zhang, P.~Li, G.~Sun, Y.~Guan, B.~Xiao, and J.~Cong, ``Optimizing fpga-based
  accelerator design for deep convolutional neural networks,'' in
  \emph{Proceedings of the 2015 ACM/SIGDA International Symposium on
  Field-Programmable Gate Arrays}.\hskip 1em plus 0.5em minus 0.4em\relax ACM,
  2015.

\bibitem{timeloop}
Y.~S. S. Y.-H. C. V. A. Y. A. M. R. V. B. K. S. W.~K. Angshuman~Parashar,
  Priyanka~Raina and J.~Emer, ``Timeloop: A systematic approach to dnn
  accelerator evaluation,'' in \emph{{ISPASS}}, 2019.

\bibitem{wu2019accelergy}
{Wu, Yannan N. and Emer, Joel S. and Sze, Vivienne}, ``{Accelergy: An
  Architecture-Level Energy Estimation Methodology for Accelerator Designs},''
  in \emph{{IEEE/ACM International Conference On Computer Aided Design
  (ICCAD)}}, {2019}.

\bibitem{lu2017flexflow}
W.~Lu, G.~Yan, J.~Li, S.~Gong, Y.~Han, and X.~Li, ``Flexflow: A flexible
  dataflow accelerator architecture for convolutional neural networks,'' in
  \emph{HPCA}, 2017.

\bibitem{kwon2018maeri}
H.~Kwon, A.~Samajdar, and T.~Krishna, ``Maeri: Enabling flexible dataflow
  mapping over dnn accelerators via reconfigurable interconnects,'' in
  \emph{ASPLOS}, 2018.

\bibitem{qin2019}
H.~K. V. N.-S. S. D. D. B. K. T.~K. Eric~Qin, Ananda~Samajdar, ``{SIGMA: A
  Sparse and Irregular GEMM Accelerator with Flexible Interconnects for DNN
  Training},'' in \emph{{HPCA}}, 2019.

\bibitem{zhu2019micro}
M.~Zhu, T.~Zhang, Z.~Gu, and Y.~Xie, ``Sparse tensor core: Algorithm and
  hardware co-design for vector-wise sparse neural networks on modern gpus,''
  in \emph{Proceedings of the 52Nd Annual IEEE/ACM International Symposium on
  Microarchitecture}, 2019.

\bibitem{wang2019hpca}
{Xiaowei Wang and Jiecao Yu and Charles Augustine and Ravi Iyer and Reetuparna
  Das}, ``{Bit Prudent In-Cache Acceleration of Deep Convolutional Neural
  Networks},'' in \emph{{HPCA}}, 2019.

\bibitem{scnn}
A.~Parashar, M.~Rhu, A.~Mukkara, A.~Puglielli, R.~Venkatesan, B.~Khailany,
  J.~S. Emer, S.~W. Keckler, and W.~J. Dally, ``{SCNN:} an accelerator for
  compressed-sparse convolutional neural networks,'' in \emph{{ISCA}}, 2017.

\bibitem{eie}
S.~Han, X.~Liu, H.~Mao, J.~Pu, A.~Pedram, M.~Horowitz, and W.~Dally, ``{EIE}:
  Efficient inference engine on compressed deep neural network,'' in
  \emph{ISCA}, 2016.

\bibitem{FPDNN}
Y.~Guan, H.~Liang, N.~Xu, W.~Wang, S.~Shi, X.~Chen, G.~Sun, W.~Zhang, and
  J.~Cong, ``{FP-DNN:} an automated framework for mapping deep neural networks
  onto {FPGAs} with {RTL-HLS} hybrid templates,'' in \emph{FCCM}, 2017.

\bibitem{DLAU}
C.~Wang, L.~Gong, Q.~Yu, X.~Li, Y.~Xie, and X.~Zhou, ``{DLAU:} {A} scalable
  deep learning accelerator unit on {FPGA},'' \emph{{IEEE} Trans. on {CAD} of
  Integrated Circuits and Systems}, vol.~36, 2017.

\bibitem{MemManagementFPGA}
X.~Wei, Y.~Liang, and J.~Cong, ``Overcoming data transfer bottlenecks in
  fpga-based {DNN} accelerators via layer conscious memory management,'' in
  \emph{DAC}, 2019.

\bibitem{shafiee2016isaac}
A.~Shafiee, A.~Nag, N.~Muralimanohar, R.~Balasubramonian, J.~P. Strachan,
  M.~Hu, R.~S. Williams, and V.~Srikumar, ``Isaac: A convolutional neural
  network accelerator with in-situ analog arithmetic in crossbars,'' in
  \emph{ISCA}, 2016.

\bibitem{wang2018dac}
P.~Wang, Y.~Ji, C.~Hong, Y.~Lyu, D.~Wang, and Y.~Xie, ``{SNrram: An Efficient
  Sparse Neural Network Computation Architecture Based on Resistive
  Random-Access Memory},'' in \emph{{DAC}}, 2018.

\bibitem{simba}
Y.~S. Shao, J.~Clemons, R.~Venkatesan, B.~Zimmer, M.~Fojtik, N.~Jiang,
  B.~Keller, A.~Klinefelter, N.~R. Pinckney, P.~Raina, S.~G. Tell, Y.~Zhang,
  W.~J. Dally, J.~S. Emer, C.~T. Gray, B.~Khailany, and S.~W. Keckler, ``Simba:
  Scaling deep-learning inference with multi-chip-module-based architecture,''
  in \emph{MICRO}, 2019.

\bibitem{chungfpga2011}
E.~S. Chung, J.~C. Hoe, and K.~Mai, ``Coram: An in-fabric memory architecture
  for fpga-based computing,'' in \emph{FPGA}, 2011.

\bibitem{ulayer2019}
Y.~Kim, J.~Kim, D.~Chae, D.~Kim, and J.~Kim, ``ulayer: Low latency on-device
  inference using cooperative single-layer acceleration and processor-friendly
  quantization,'' in \emph{Proceedings of the Fourteenth EuroSys Conference
  2019}, 2019.

\bibitem{boroumand2018}
A.~Boroumand, S.~Ghose, Y.~Kim, R.~Ausavarungnirun, E.~Shiu, R.~Thakur, D.~Kim,
  A.~Kuusela, A.~Knies, P.~Ranganathan, and O.~Mutlu, ``{Google Workloads for
  Consumer Devices: Mitigating Data Movement Bottlenecks},'' in
  \emph{{ASPLOS}}, 2018.

\bibitem{zhu2018vision}
\BIBentryALTinterwordspacing
Y.~Zhu, A.~Samajdar, M.~Mattina, and P.~Whatmough, ``Euphrates: Algorithm-soc
  co-design for low-power mobile continuous vision,'' in \emph{Proceedings of
  the 45th Annual International Symposium on Computer Architecture}, ser. ISCA
  '18.\hskip 1em plus 0.5em minus 0.4em\relax Piscataway, NJ, USA: IEEE Press,
  2018. [Online]. Available: \url{https://doi.org/10.1109/ISCA.2018.00052}
\BIBentrySTDinterwordspacing

\bibitem{eva2}
M.~Buckler, P.~Bedoukian, S.~Jayasuriya, and A.~Sampson, ``Exploiting temporal
  redundancy in live computer vision,'' \emph{ISCA}, 2018.

\bibitem{asv}
Y.~Feng, P.~N. Whatmough, and Y.~Zhu, ``{ASV:} accelerated stereo vision
  system,'' in \emph{{MICRO}}, 2019.

\bibitem{sadri_fpgaworld2013}
M.~Sadri, C.~Weis, N.~Wehn, and L.~Benini, ``Energy and performance exploration
  of accelerator coherency port using xilinx zynq,'' in \emph{FPGAWorld}, 2013.

\bibitem{moreau2015snnap}
T.~Moreau, M.~Wyse, J.~Nelson, A.~Sampson, H.~Esmaeilzadeh, L.~Ceze, and
  M.~Oskin, ``Snnap: Approximate computing on programmable socs via neural
  acceleration,'' in \emph{HPCA}, 2015.

\end{thebibliography}

\end{document}